\documentclass[lettersize,journal]{IEEEtran}
\usepackage{amsmath,amsfonts}
\usepackage{algorithmic}
\usepackage{algorithm}
\usepackage{array}
\usepackage[caption=false,font=normalsize,labelfont=sf,textfont=sf]{subfig}
\usepackage{textcomp}
\usepackage{stfloats}
\usepackage{url}
\usepackage{verbatim}
\usepackage{graphicx}
\usepackage{cite}

\usepackage{multirow}
\usepackage{hyperref}
\usepackage{amsmath}
\usepackage{bm}

\begin{document}

\title{CV2X-LOCA: Roadside Unit-Enabled Cooperative Localization Framework for Autonomous Vehicles}

\author{Zilin Huang, Sikai Chen*, Yuzhuang Pian, Zihao Sheng, Soyoung Ahn, and David A. Noyce 
\thanks{This work was supported in part by ... (Corresponding author: Sikai Chen.)}
\thanks{Zilin Huang, Sikai Chen, Zihao Sheng, Soyoung Ahn, and David A. Noyce are with the College of Engineering, University of Wisconsin-Madison, Madison, WI, 53706, USA (E-mails: \{zilin.huang, sikai.chen, Zihao sheng, sue.ahn, danoyce\}@wisc.edu).
Yuzhuang Pian is with the Department of Computer Science and Technology, Sun Yat-sen University, Guangzhou 510006, China (E-mails: pianyzh@mail2.sysu.edu.cn)}}

\markboth{Journal of \LaTeX\ Class Files,~Vol.~14, No.~8, August~2021}%
{Shell \MakeLowercase{\textit{et al.}}: A Sample Article Using IEEEtran.cls for IEEE Journals}


\maketitle

\begin{abstract}
An accurate and robust localization system is crucial for autonomous vehicles (AVs) to enable safe driving in urban scenes. While existing global navigation satellite system (GNSS)-based methods are effective at locating vehicles in open-sky regions, achieving high-accuracy positioning in urban canyons such as lower layers of multi-layer bridges, streets beside tall buildings, tunnels, etc., remains a challenge. In this paper, we investigate the potential of cellular-vehicle-to-everything (C-V2X) wireless communications in improving the localization performance of AVs under GNSS-denied environments. Specifically, we propose the first roadside unit (RSU)-enabled cooperative localization framework, namely CV2X-LOCA, that only uses C-V2X channel state information to achieve lane-level positioning accuracy. CV2X-LOCA consists of four key parts: data processing module, coarse positioning module, environment parameter correcting module, and vehicle trajectory filtering module. These modules jointly handle challenges present in dynamic C-V2X networks. Extensive simulation and field experiments show that CV2X-LOCA achieves state-of-the-art performance for vehicle localization even under noisy conditions with high-speed movement and sparse RSUs coverage environments. The study results also provide insights into future investment decisions for transportation agencies regarding deploying RSUs cost-effectively.
\end{abstract}


\section{Introduction}
\begin{figure*}[!t]
\centering
\subfloat[]{\includegraphics[width=0.45\textwidth,height=2.2in]{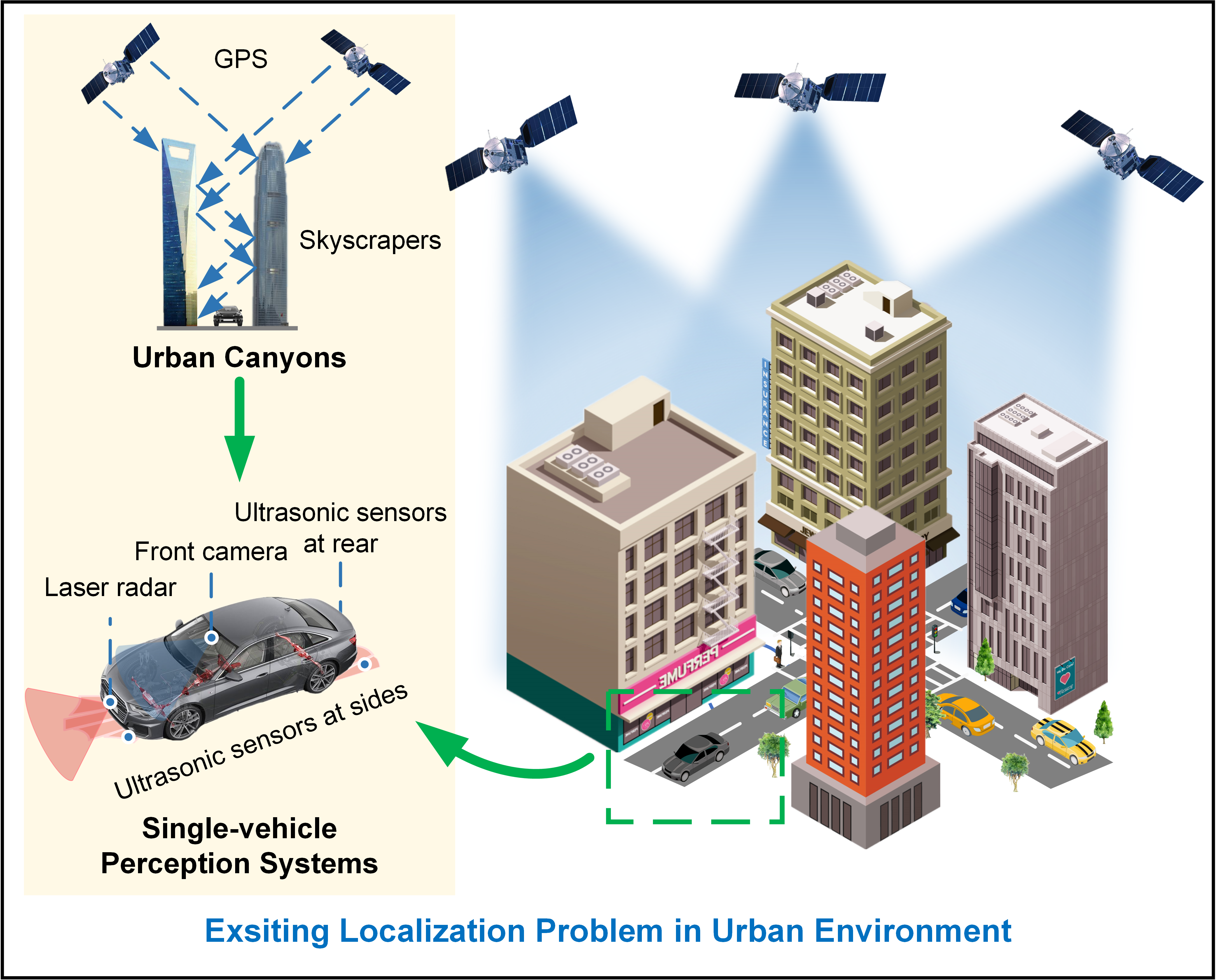}
\label{fig1_1}}
\subfloat[]{\includegraphics[width=0.52\textwidth,height=2.2in]{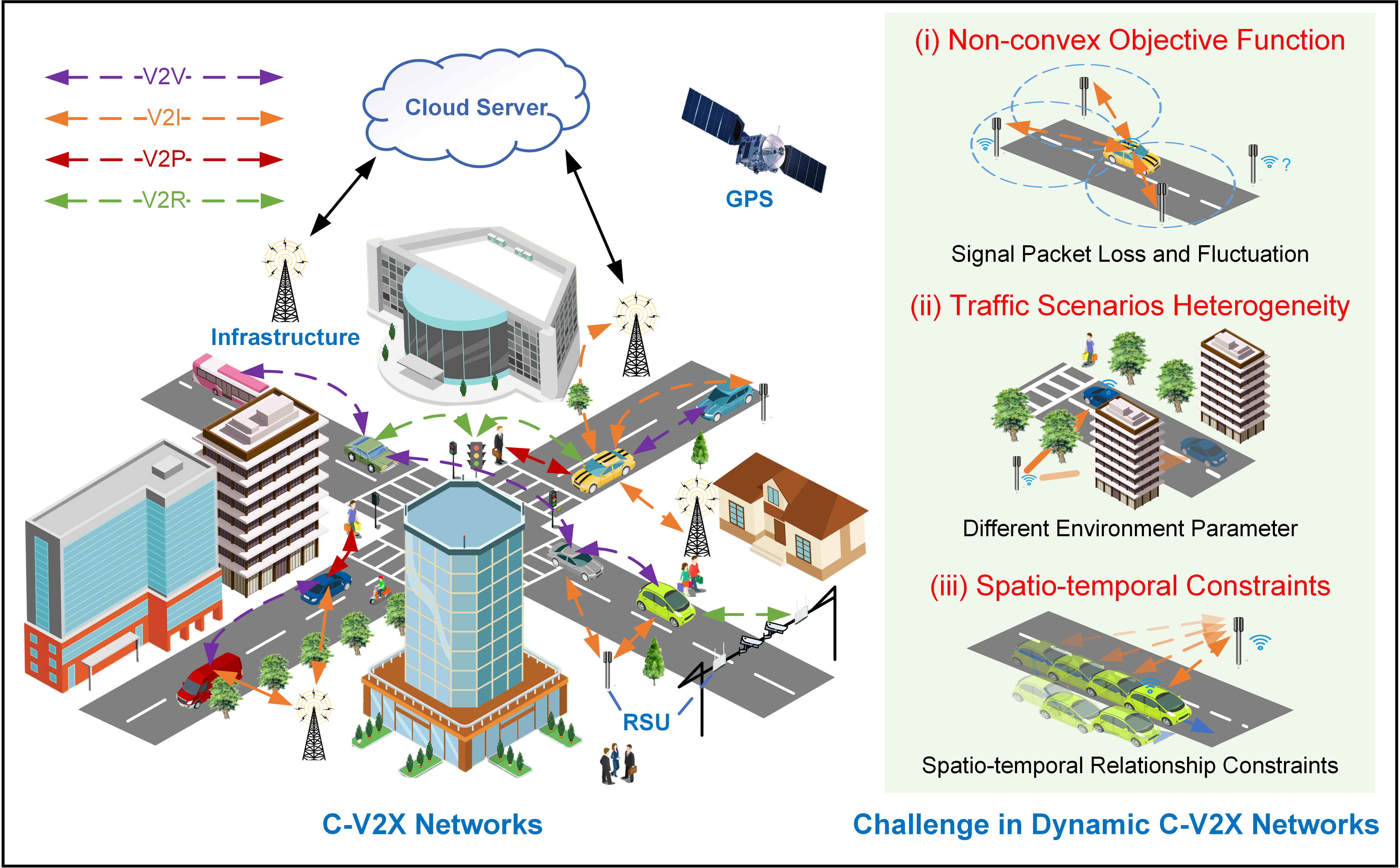}
\label{fig1_2}}
\caption{An illustration of AVs driving in an urban canyon scenario. (a) a motivating example for our proposed CV2X-LOCA framework, which aims to address the existing localization problem faced by traditional single-vehicle perception systems in GNSS-denied environments. (b) essential components of C-V2X networks, and challenges associated with locating vehicles based on channel state information in dynamic C-V2X networks.}
\label{fig1}
\end{figure*}

\IEEEPARstart{A}{utonomous} vehicles (AVs) technologies have garnered significant attention in recent years due to their potential in promoting safe driving, alleviating traffic congestion, and mitigating energy consumption \cite{feng2023nature, eskandarian2019research, matin2022impacts}. The Society of Automotive Engineers (SAE) defines six levels of automation for AVs, ranging from 0 (fully manual) to 5 (fully autonomous) \cite{drivinglevels}. To achieve a higher level of automation, it is crucial to ensure real-time, accurate and robust vehicle positioning by perceiving the complex driving environment \cite{lu2021real}. Conventionally, the positioning methods used for land vehicles employ an integrated system that combines a global navigation satellite system (GNSS) and an inertial navigation system (INS) \cite{jing2022integrity}. However, GNSS signal outage is a significant problem in this integrated systems, especially in heavily urbanized areas \cite{chiang2020performance, zhu2018gnss, ma2017radar, kim2022tunnel, panev2018road, wang2022pavement, qin2017vehicles}. This issue arises because GNSS signals can sometimes be weak, highly distorted, or even denied in ``urban canyons", such as lower layers of multi-layer bridges, streets beside tall buildings, tunnels, and areas with dense vegetation or foliage. Fig. \ref{fig1} (a) shows an example, where a vehicle is driving in an urban canyon surrounded by buildings that block GNSS signals, rendering insufficient satellite signals for vehicle localization (which needs at least three satellites). Moreover, the navigation performance of inertial measurement units (IMUs) in INS systems deteriorates over time, resulting in degradation or failure of the positioning accuracy using GNSS-based methods \cite{zhu2018gnss}. Therefore, it is necessary to find alternative methods to achieve better positioning accuracy for safe AV operations in GNSS-denied environments.

One approach to address the GNSS-denied problem is map matching with sufficiently precise maps \cite{rohani2015novel}. This technique constrains vehicle location to the roadway, correcting erroneous estimates that appear outside of it. However, it requires a map database and cannot guarantee lane-level accuracy while GNSS signals are unavailable. Another solution is simultaneous localization and mapping (SLAM), which builds an environment model while locating vehicles \cite{lu2021real}. SLAM algorithms focus on abstract data from on-board perception sensors, such as Lidar \cite{kim2022tunnel}, Radar \cite{ma2017radar}, Camera \cite{panev2018road}, or their combination \cite{lee2020fail}. Reliable metropolitan high definition (HD) map construction and robust environmental perception are two main challenges for SLAM, especially in adverse conditions like snow or darkness. To achieve accurate positioning, a multi-sensor fusion scheme has been proposed \cite{chiang2020performance}, but increases complexity significantly. Moreover, perceiving the environment solely through on-board perception sensors mounted on one ego-vehicle has limitations such as occlusion and limited sensing range \cite{xu2022v2x}. With the growing use of vehicle-to-everything (V2X) wireless communication technology, V2X signals have begun to be adopted as positioning and navigation components for AVs during GPS outages \cite{adegoke2019infrastructure, liu2022research, del2019network, yang2020multi,li2018rse,linrssi}. 

The V2X signal-based positioning method can achieve global positioning of vehicles when GNSS signals are weak or unavailable. Additionally, it can solve non-line-of-sight (NLOS) issues related to on-board perception sensors at a relatively low cost\cite{lu2021real,adegoke2019infrastructure,ko2021v2x}. The fundamental principle of this localization approach involves measuring physical quantities of V2X channel state information, such as time of arrival (TOA), time difference of arrival (TDOA), angle of arrival (AOA), and received signal strength (RSS) \cite{saeed2018localization, brambilla2019augmenting, zhuang2022novel, ko2021v2x}. There are two main contenders for V2X technologies: dedicated short-range communication (DSRC) and cellular vehicle-to-everything (C-V2X) \cite{moradi2023dsrc, maglogiannis2021experimental}. Currently, C-V2X has gained significant attention from industry and academia, as a next-generation V2X wireless communication technology that aims to improve road safety and traffic efficiency \cite{liu2021highly,kutila2019c}. Although C-V2X technology has the potential to revolutionize the mobile industry, its ability to enhance vehicle positioning accuracy has not been fully explored \cite{ko2021v2x, liu2021highly}. Three challenges remain in locating vehicles within dynamic C-V2X networks, as illustrated in Fig. \ref{fig1} (b): (i) non-convex objective function, (ii) traffic scenarios heterogeneity, and (iii) spatio-temporal constraints. Section \ref{II-C} provides a detailed discussion of these challenges.

Table \ref{Tab1} summarizes existing sate-of-the-art methods developed to improve V2X-based positioning accuracy in response to challenges (i) through (iii). To address challenge (i), Zhu \textit{et al.} \cite{jianyong2014rssi}, Saeed \textit{et al.} \cite{saeed2018localization}, and Ma \textit{et al.} \cite{ma2019efficient} used Taylor series expansions, linear expansions, and weighted position information to approximate the original nonlinear equation system as a set of linear equations relating to the vehicle's position. However, this least squares (LS)-based framework is ineffective when measurement noise variance is significant. Other researchers \cite{chen2016vehicle, magowe2019closed} attempted to use weighted centroid localization (WCL)-based framework, which is simpler but less effective than the maximum likelihood (ML) estimator. The semi-definite programming (SDP) framework has gained interest in wireless communication field for improving positioning accuracy \cite{zheng2020accurate, wang2018cooperative, zhang2021localization,zou2021rss}; however it is not suitable for dynamic road environments that pose challenges (ii) and (iii). Page \textit{et al.} \cite {page2019enhanced} and Zhang \textit{et al.} \cite {zhang2021localization} integrated extended Kalman filter (EKF) and Kalman filter (KF) into the LS- and SDP-based frameworks respectively, to simultaneously tackle challenges (i) and (iii). Nevertheless these solutions do not consider challenge (ii), which requires non-linear problem-solving techniques, since KF and EKF are only suitable for linear problems. Huang \textit{et al.} \cite {huang2020multi} and Jondhale \textit{et al.} \cite{jondhale2018kalman } improved signal-distance models using curve fitting(CF) and generalized regression neural network (GRNN) techniques, respectively, and incorporated an unscented Kalman filter (UKF) to enhance positioning accuracy. Although these improved signal-distance models can achieve better localization performance in different road environments, they do not address challenge (i).
 
\begin{table}[]
\caption{Existing V2X-Based Localization Methods.}
\begin{tabular}{ccccc}
\hline
\multirow{2}{*}{\textbf{Methods}} & \multirow{2}{*}{\textbf{Category}}     & \multicolumn{3}{c}{\textbf{Challenges}}     \\ \cline{3-5} 
                                  &                                        & \textbf{(i)} & \textbf{(ii)} & \textbf{(iii)} \\ \hline
ML                         & ML estimator in Eq. (\ref{Eq2})                  &             &              &              
     \\ \hline
LS-TS \cite{jianyong2014rssi}                            & \multirow{3}{*}{LS-based   framework}  & \checkmark           &              &              \\
LLS \cite{saeed2018localization}                              &                                        & \checkmark            &              &              \\
WLLS \cite{ma2019efficient}                             &                                        & \checkmark            &              &              \\ \hline
WCL \cite{chen2016vehicle}                              & \multirow{2}{*}{WCL-based   framework} & \checkmark            &              &              \\
WCL-TS \cite{magowe2019closed}                           &                                        & \checkmark            &              &              \\ \hline
SDP-LSRE \cite{wang2018cooperative}                         & SDP-based framework                    & \checkmark            &              &              \\ \hline
LS-EKF \cite{page2019enhanced}                           & LS+EKF-based framework                 & \checkmark            &              & \checkmark            \\ \hline
SDP-ML-KF \cite{zhang2021localization}                        & SDP+KF-based framework                 & \checkmark            &              & \checkmark            \\ \hline
GRNN-UKF  \cite{jondhale2018kalman}                        & GRNN+UKF-based framework               &              & \checkmark            & \checkmark            \\ \hline
CF-LS-UKF \cite{huang2020multi}                        & CF+UKF-based framework                 &              & \checkmark            & \checkmark            \\ \hline
CV2X-LOCA (Ours)                    & CV2X-LOCA-based framework               & \checkmark            & \checkmark            & \checkmark            \\ \hline
\end{tabular}
NOTE: “\checkmark” indicates that the method focuses on solving this challenge. 
\label{Tab1}
\end{table}

As far as the author knows, there is currently no established industry standard or mature solution for utilizing C-V2X channel state information to improve AVs' positioning accuracy, due to aforementioned challenges. This paper aims to develop a framework for AVs localization that can address these challenges and achieve lane-level positioning accuracy in GNSS-denied environments. The results of this research can also provide valuable insights to transportation agencies to make cost-effective decisions for deploying roadside units (RSUs) (e.g., optimal deployment spacing). The major contributions of this work are as follows:

(1) We propose the first unified RSU-enabled cooperative localization framework, namely CV2X-LOCA\footnote{For those interested in exploring further, demo code is available at: \url{https://github.com/julianwong-source/CV2X-LOCA}.}, which can jointly handle challenges present in dynamic C-V2X networks and achieve lane-level positioning accuracy under GNSS-denied environments. The CV2X-LOCA relies solely on C-V2X channel state information and does not require expensive on-board perception sensors. Extensive simulations and field experiments demonstrate that CV2X-LOCA achieves state-of-the-art localization performance in various traffic scenarios, even under noisy conditions with high-speed movement and sparse RSUs coverage environments.

(2) We develop a coarse positioning module designed for locating vehicles in dynamic C-V2X networks. This module can solve the problem of non-convex and highly nonlinear objective function in traditional ML estimator for V2X signal-based vehicle localization. We first derive a non-convex estimator that approximates the ML estimator but has no logarithm in the residual, and then apply semi-definite relaxation techniques to develop a convex estimator that can be efficiently solved. 

(3) We design an environment parameter correcting module that can address the problem of signal propagation model changes in a dynamic C-V2X network when locating moving vehicles. Specifically, we introduce a novel concept, called cooperative roadside units (C-RSUs), which can capture the heterogeneity of traffic scenarios. This correction mechanism is very promising for V2X signal-based localization tasks as it can be easily integrated to existing well-established studies.

(4) We present a vehicle trajectory filtering module that utilizes prior traffic information, i.e., vehicle motion features, to improve positioning accuracy. In particular, a trajectory-based technique of the UKF is leveraged to filter coarse positioning points estimated by the coarse positioning module after correction for environmental parameters. The outlier estimates can be further smoothed by incorporating the filtering mechanism, thus enhancing the positioning accuracy.

The paper is structured as follows: Section \ref{II} covers the preliminaries, while Section \ref{III} introduces the CV2X-LOCA framework and provides detailed information on each module. Sections \ref{IV} and \ref{V} report simulation and field experiment results, respectively. Computational complexity is discussed in Section \ref{VI}, and finally, Section \ref{VII} concludes the paper.

\section{Preliminaries} \label{II}

\subsection{RSU-Enabled Vehicular Network} \label{II-A}
\begin{figure}[!t]
\centering
\includegraphics[width=3.48in,height=1.85in]{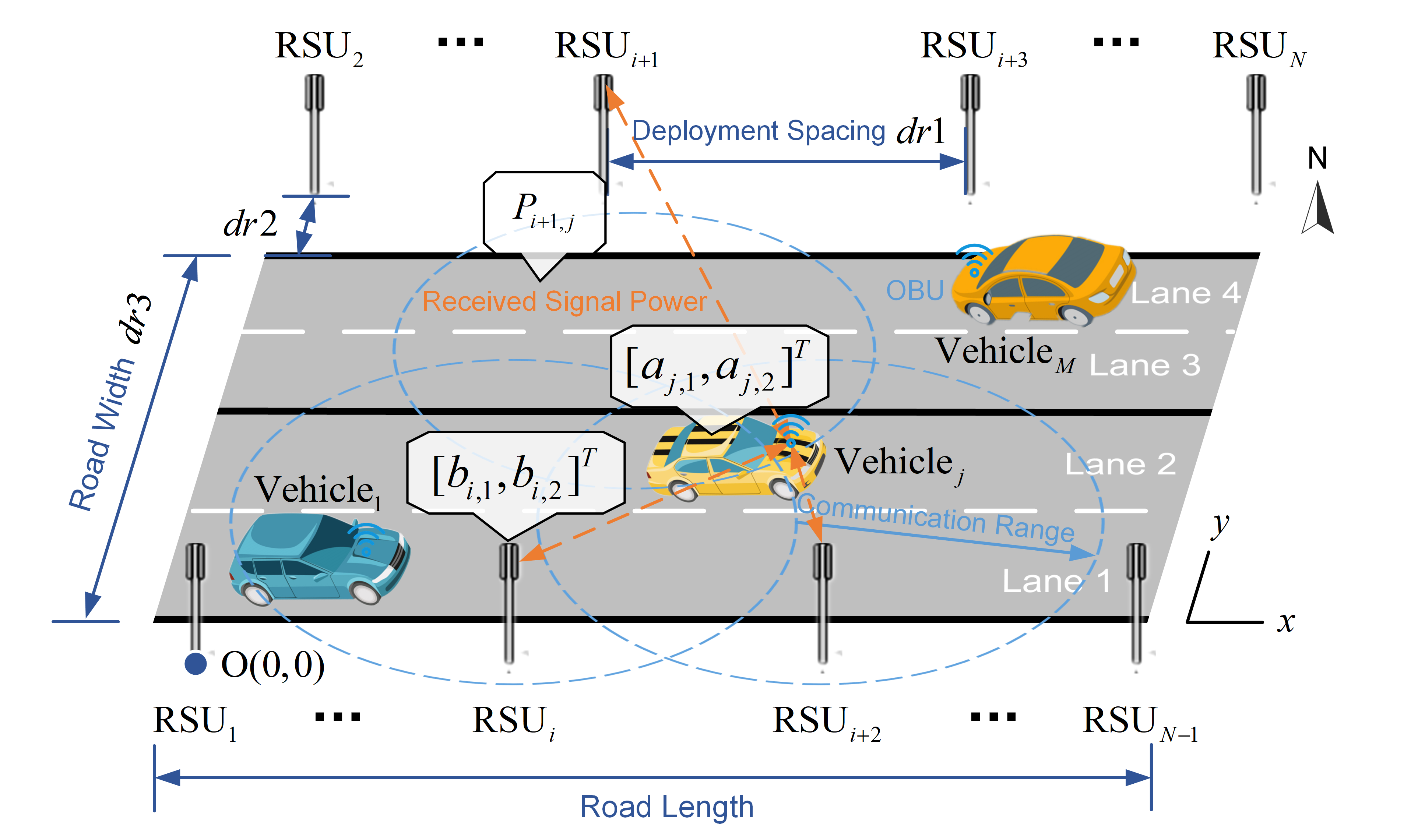}
\caption{Illustration of the vehicular network considered in this paper.}
\label{fig2}
\end{figure}

In this paper, we consider a two-way four-lane vehicular network, consisting of two types of connected devices: on-board units (OBUs) installed in vehicles and RSUs deployed on roadsides, as illustrated in Fig. \ref{fig2}. The data used for positioning a vehicle is RSS, parsed from a built-in received-signal-strength-indicator (RSSI), which can be easily extracted from a standard message set. The RSSI measures strength of signal from the RSU, which is then stored temporarily at the OBU device and can later be uploaded to cloud servers \cite{maglogiannis2021experimental, soto2022survey}. Without loss of generality, we assume that the position of a vehicle can be described using two-dimensional (2D) coordinates. The X and Y-axis of the local coordinate system are assumed to be parallel and perpendicular to the lanes, respectively. The unknown coordinates of the $j$th vehicle are denoted as $\boldsymbol{\theta}_{j}=\left[a_{j, 1}, a_{j, 2}\right]^{T}\left(\boldsymbol{\theta}_{j} \in \mathbb{R}^{2}, j=1, \ldots, M\right)$, and the known coordinates of the $i$th RSU are denoted as $\boldsymbol{\phi_{i}}=\left[b_{i, 1}, b_{i, 2}\right]^{1}\left(\bm{\phi_{i}} \in \mathbb{R}^{2}, i=1, \ldots, N\right)$. Here, $M$ represents the total number of vehicles, while $N$ denotes the number of RSUs that each vehicle can communicate with, which is determined by factors such as RSUs' communication range and transmitting power. Each deployment's coordinate origin is marked as $O$ and shown in the corresponding plot at (0, 0). Additionally, $dr_1$ denotes the deployment distance between two RSUs, $dr_2$ represents the distance between an RSU and the road edge, while $dr_3$ represents the width of the road.

\subsection{Signal Propagation Model}
Accurately determining positioning relies heavily on signal propagation modeling. Various wireless communication channel models are available, including the Okumura-Hata model, Lee model, standard indoor model, attenuation factor model, linear loss model, and logarithmic path loss model \cite{liu2022research}. Among them, the logarithmic loss model holds that path loss of signal propagation is logarithmically related to the increase in distance, and can be considered as a simple and accurate description of signal propagation characteristics in traffic scenarios \cite{liu2022research, li2018rse, saeed2018localization,mafakheri2021optimizations,miao2022does}. 

The received signal power (in dBm) at the $j$th vehicle transmitted from the $i$th RSU, denoted as $P_{i, j}$, is related to their distance using the logarithmic path loss model
\begin{equation}
P_{i, j}=P_{0}+10 \gamma \log _{10} \frac{\left\|\bm{\theta_{j}}-\bm{\phi_{i}}\right\|}{d_{0}}+m_{i, j}
\label{Eq1}
\end{equation}
where $P_{0}$ denotes the received signal power at a reference distance $d_{0}$ ($\left\|\boldsymbol{\theta_{j}}-\boldsymbol{\phi_{i}}\right\| \geq d_{0}$), $\left\|\bm{\theta_{j}}-\bm{\phi_{i}}\right\|$ means the Euclidean distance between the $i$th RSU at the $j$th vehicle, and noise $m_{i, j}$ represents the log-normal shadowing effect in multipath environments, respectively. Typically, $m_{i, j}$ is modeled as a zero mean Gaussian random variable with shadowing standard deviation $\sigma_{i, j}^2$, i.e., $m_{i, j} \sim \mathcal{N}\left(0, \sigma_{i j}^2\right)$ \cite{gholami2011positioning}. The path-loss exponent (PLE), $\gamma$, which ranges from 2 to 6 \cite{ouyang2010received}, is also known as an environmental parameter (influenced by surrounding environment). Given RSS measurements and environmental parameters, we estimate the parametric localization problem for the $j$th vehicle using Eq. (\ref{Eq1}) to determine its location $\boldsymbol{\theta}_{j}$.

\subsection{Problem Statement} \label{II-C}
As can be seen, the model is nonlinearly dependent on the location of the $j$th vehicle. Here, $f\left(\bm{\theta_{j}}-\bm{\phi_{i}}\right)=P_{0}+10 \gamma \log _{10} \frac{\left\|\boldsymbol{\theta}_{j}-\bm{\phi_{i}}\right\|}{d_{0}}$. Suppose measurement errors are independent and identically distributed (i.i.d.). The ML estimator for RSS-based localization problem can be derived as \cite{gholami2011positioning}
\begin{equation}
\hat{\boldsymbol{\theta}}_{j}=\arg \min _{\theta_{j}} \sum_{i=1}^{N}\left(-\left(P_{i, j}-P_{0}\right)+10 \gamma \log _{10} \frac{\left\|\bm{\theta_{j}}-\bm{\phi_{i}}\right\|}{d_{0}}\right)^{2}
\label{Eq2}
\end{equation}

However, by the above mentioned, there are three challenges that hinder the direct use of Eq. (\ref{Eq2}) for vehicle localization in dynamic C-V2X networks. Firstly, as it is observed, the original ML estimator is highly non-convex due to its domain $\left\{\boldsymbol{\theta}_{j} \mid \boldsymbol{\theta}_{j} \neq \boldsymbol{\phi_{i}}\right\}$ is not continuous \cite{ouyang2010received}. Also, the C-V2X signal is more prone to packet loss in traffic scenarios compared to static sensor networks \cite{maglogiannis2021experimental}. Furthermore, Eq. (\ref{Eq2}) contains $\log _{10}\left\|\boldsymbol{\theta}_{j}-\boldsymbol{\phi_{i}}\right\|$, which is neither convex nor concave in the objective function. Although it is possible to restrict $\boldsymbol{\theta}_{j}$ within a convex domain (as verified by examining the Hessian of $\log _{10}\left\|\boldsymbol{\theta}_{j}-\boldsymbol{\phi_{i}}\right\|$, which is neither positive semi-definite nor negative semi-definite), this approach is impractical due to signal fluctuations caused by rapidly moving targets in traffic scenarios \cite{huang2019novel}. Consequently, finding and confirming a global minimum solution becomes challenging. A convex estimator would be highly desirable, but the presence of $\log _{10}\left\|\boldsymbol{\theta}_{j}-\boldsymbol{\phi_{i}}\right\|$ makes such an idea troublesome. 

Secondly, in dynamic C-V2X networks, vehicles traverse various road environments such as urban forests and skyscrapers, resulting in different environmental parameter in Eq. (\ref{Eq2}). Meanwhile, wireless signal attenuation can be affected by various factors, including multi-path fading, diffraction, reflection, and weather conditions \cite{huang2020multi}. 
 Nguyen \textit{et al.} \cite{nguyen2022cellular} revealed that large vehicle shadowing also impacts on V2V communication in Mode 4 C-V2X networks. The localization system would be become vulnerable if the environmental parameter is not estimated correctly. Thirdly, Eq. (\ref{Eq2}) is derived from a static sensor network, which assumes that the target moves randomly and its location can only be calculated based on single-snapshot observations at each observation time \cite{wang2018cooperative}. In traffic scenarios, there is prior information available. For instance, the position of the target at the last moment has a high degree of spatial and temporal relationship with its location at the next moment. Additionally, longitudinal and lateral motion features can be defined and modeled using mathematical representations \cite{page2019enhanced,jo2016tracking}.

 \begin{figure*}[!t]
\centering
\includegraphics[width=0.99\textwidth,height=2.9in]{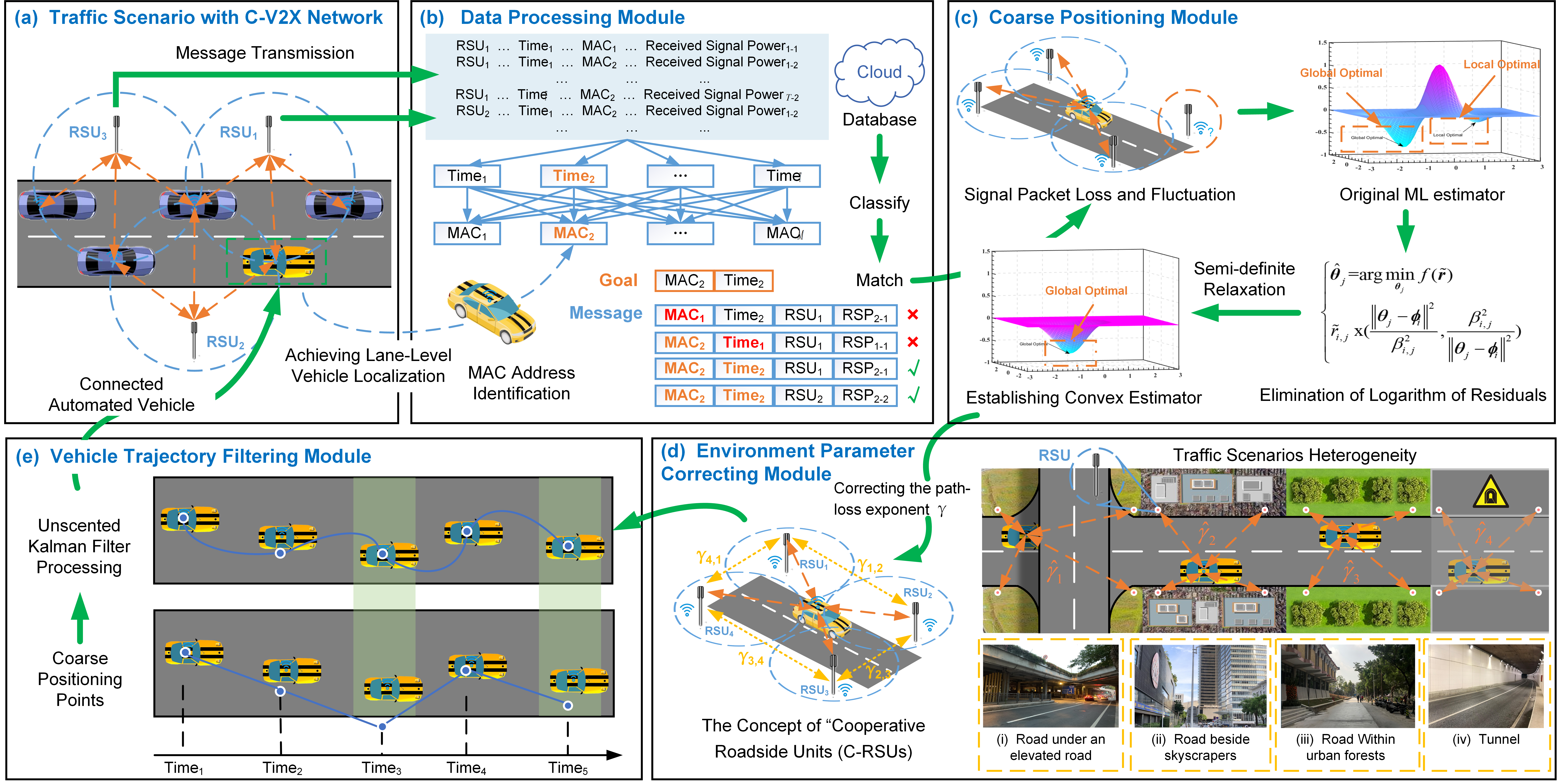}
\caption{The CV2X-LOCA framework architecture. (a) traffic scenario with C-V2X network. (b) data processing module presented in \ref{III-B}. (c) coarse positioning module proved in \ref{III-C}. (d) environment parameter correcting module proposed in \ref{III-D}. (e) vehicle trajectory filtering module illustrated in \ref{III-E}.}
\label{fig3}
\end{figure*}

\section{Roadside Unit-Enabled Cooperative Localization Framework} \label{III}
\subsection{CV2X-LOCA Overview} \label{III-A}
Our goal is to develop a reliable C-V2X-based vehicle localization framework that can overcome the challenges mentioned above and achieve lane-level positioning accuracy even when GNSS signals are unavailable. The architecture of proposed localization framework, namely CV2X-LOCA, is shown in Fig. \ref{fig3}. Specifically, CV2X-LOCA is comprised of four main components: (1) a data processing module, (2) a coarse positioning module, (3) an environment parameter correcting module, and (4) a vehicle trajectory filtering module. These modules work together to address the challenges present in dynamic C-V2X networks. 

As depicted in Fig. \ref{fig3} (a), using C-V2X communications, the OBU mounted on a vehicle sends messages periodically to all surrounding network nodes such as vehicles, infrastructure and pedestrians to provide and update information about itself \cite{zhuang2022novel,soto2022survey}. For instance, the vehicle can broadcast sensor data like GNSS positions through Cooperative Awareness Message (CAM), which is defined by ETSI and usually used in Europe \cite{etsi2011intelligent}, or Basic Safety Message (BSM), which is defined by SAE International and typically used in the U.S. \cite{sae2022V2X}. Additionally, RSUs can be considered communicable infrastructures with very accurate position information but unknown transmit power. We assume that there may be noise interference during transmission of C-V2X signal while signal packets may also fluctuate or go missing due to high mobility of vehicles and other environmental factors.

\subsection{Data Processing Module} \label{III-B}
We propose a data processing module, illustrated in Fig. \ref{fig3} (b), to match data from different OBUs. The matrix $\boldsymbol{G}_{i}$ represents the signal dataset collected by all vehicles from the $i$th RSU during the detection period $T$, sorted in ascending time order. The data samples are as follows $\boldsymbol{G}_{i}$=\{\textit{00:16:EA:AE:3C:30, 7/10/2021 10:20:30, -68; ...}\}, where "\textit{00:16:EA:AE:3C:30}" denotes the unique media access control address (MAC address) for each vehicle ($j$th vehicle), "\textit{7/10/2021 10:20:30}" denotes the timestamp, and "\textit{-68}" denotes received signal power values between all vehicles and the $i$th RSU. Similarly, we consider the matrix $\boldsymbol{G}_{j}$ as the signal dataset collected by the $j$th vehicle from all RSUs during detection period $T$, sorted in ascending time order. The data samples are as follows $\boldsymbol{G}_{i}$=\{\textit{7/10/2021 10:20:30, $R S U_1$,-68; ...}\}, where the "$R S U_1$" denotes the absolute coordinate positions of the first RSU. By matching these datasets using unique MAC addresses and timestamps as identifiers, we can obtain sample data collected by all RSUs as follows
\begin{equation}
\boldsymbol{G}_{i, j}=\left[\begin{array}{c}
\mathrm{MAC}_{j},7/10/202110:20:30,\mathrm{RSU}_{1}, -68 \\
\ldots \\
\mathrm{MAC}_{j},7/10/202110:20:30,\mathrm{RSU}_{i}, -66
\end{array}\right]
\label{Eq3}
\end{equation}
where the first row represents the data collected by the first RSU from $j$th vehicle at time $t$, and so on. With the above data processing, we can match the data collected from each vehicle by all RSUs during detection period of $T$.

\subsection{Coarse Positioning Module} \label{III-C}
From the above, we should first solve the non-convex and highly nonlinear objective function of conventional ML estimator of Eq. (\ref{Eq2}). Inspired by the newly developed SDP-based framework \cite{zheng2020accurate, wang2018cooperative, zhang2021localization, zou2021rss}, this paper presents a coarse positioning module, tailored for vehicle localization in dynamic C-V2X networks, as shown in Fig. \ref{fig3} (c).

\textit{1) Eliminate the Logarithm of Residuals:} We can define a residual in Eq. (\ref{Eq2}) as \cite{ouyang2010received}
\begin{equation}
\boldsymbol{r}_{i, j}=10 \gamma \log _{10} \frac{\left\|\bm{\theta_{j}}-\bm{\phi_{i}}\right\|}{d_{0}}-\left(P_{i, j}-P_{0}\right)
\label{Eq4}
\end{equation}
where $\boldsymbol{r}_{i, j}$ is the difference between the actual value and the measurement. Then, Eq. (\ref{Eq2}) can be considered to minimize a penalty function on the residual vector $\boldsymbol{r}=\left[r_{1, j}, \ldots, r_{N, j}\right]^{T}\left(\boldsymbol{r} \in \mathbb{R}^{N}\right)$ as
\begin{equation}
\hat{\boldsymbol{\theta}}_{j}=\arg \min _{\boldsymbol{\theta}_{j}} f(\boldsymbol{r})
\label{Eq5}
\end{equation}
where $f(.)=\|\cdot\|^{2}$.

To derive a nonconvex estimator without the logarithm in the residuals, we replace $f(.)=\|\cdot\|^{2}$ in Eq. (\ref{Eq5}) with another penalty function $f(.)=\|.\|_{\infty}=\max \left|[\cdot]_{i}\right|$, i.e., the Chebyshev norm, which is also known as the $\ell_{\infty}$ norm \cite{boyd2004convex}. 

This way, Eq. (\ref{Eq2}) can be approximated as
\begin{equation}
\hat{\boldsymbol{\theta}}_{j}=\arg \min _{\boldsymbol{\theta_{j}}} \max _{i}\left|10 \gamma \log _{10} \frac{\left\|\bm{\theta_{j}}-\boldsymbol{\phi_{i}}\right\|}{d_{0}}-\left(P_{i, j}-P_{0}\right)\right|
\label{Eq6}
\end{equation}

Now, we can form an equivalent problem without a logarithm after certain manipulations \cite{boyd2004convex}. This gives
\begin{equation}
\hat{\boldsymbol{\theta}_{j}}=\arg \min _{\boldsymbol{\theta_{j}}} \max _{i} \max \left(\frac{\left\|\bm{\theta_{j}}-\boldsymbol{\phi_{i}}\right\|^{2}}{\beta_{i, j}^{2}}, \frac{\beta_{i, j}^{2}}{\left\|\boldsymbol{\theta_{j}}-\boldsymbol{\phi_{i}}\right\|^{2}}\right)
\label{Eq7}
\end{equation}
where
\begin{equation}
\beta_{i, j}^{2}=d_{0}{ }^{2} 10^{\frac{P_{i, j}-P_{0}}{5 \gamma}}
\label{Eq8}
\end{equation}

We provide a detailed derivation and proof of Eq. (\ref{Eq7}) in Appendix \ref{App1} for interested reader. Following is the proposed estimator for the CV2X-based vehicle localization problem considered in this paper, which can simply be written as
\begin{equation}
\hat{\boldsymbol{\theta}}_{j}=\arg \min _{\boldsymbol{\theta}_{j}} f(\tilde{\boldsymbol{r}})
\label{Eq9}
\end{equation}
where $\tilde{\boldsymbol{r}}=\left[\tilde{r}_{1, j}, \ldots, \tilde{r}_{N, j}\right]^{T}\left(\tilde{\boldsymbol{r}}_{i, j} \in \mathbb{R}^{N}\right)$ is the proposed residual vector, which can be equivalently written as
\begin{equation}
\tilde{r}_{i, j}=\max \left(\frac{\left\|\bm{\theta_{j}}-\boldsymbol{\phi_{i}}\right\|^{2}}{\beta_{i, j}^{2}}, \frac{\beta_{i, j}^{2}}{\left\|\bm{\theta_{j}}-\boldsymbol{\phi_{i}}\right\|^{2}}\right)
\label{Eq10}
\end{equation}
and $f(.)$ is an arbitrary convex penalty function (e.g. the $\ell_{p}$ norm family \cite{boyd2004convex}).

\textit{Remark 1 (Rationality of Estimator)}: A closer examination of Eq. (\ref{Eq8}) shows that it can be expressed as
\begin{equation}
\beta_{i, j}^{2}=10^{\frac{m_{i, j}}{5}}\left\|\boldsymbol{\theta_{j}}-\boldsymbol{\phi_{i}}\right\|^{2}
\label{Eq11}
\end{equation}

Eq. (\ref{Eq8}) and Eq. (\ref{Eq11}) demonstrate that noise in $\beta_{i, j}^{2}$ (a function of the measurement $P_{i, j}$) is a multiplicative model. That is, the noise $m_{i, j}$ is multiplicative $\left\|\boldsymbol{\theta_{j}}-\boldsymbol{\phi_{i}}\right\|^{2}$ \cite{ouyang2010received}. Notably, it can also be represented as an additive model \cite{wang2018cooperative}. The residual $\tilde{r}_{i, j}$ in Eq. (\ref{Eq10}) is also consistent with such a multiplicative noise model, since it contains only the ratio $\left\|\boldsymbol{\theta_{j}}-\boldsymbol{\phi_{i}}\right\|^{2} / \beta_{i, j}^{2}$ and its inverse.

\textit{Remark 2 (Equivalence of Estimator)}: With $f(.)$ in Eq. (\ref{Eq9}) being the Chebyshev norm, Eq. (\ref{Eq9}) is just Eq. (\ref{Eq7}). As shown in Appendix \ref{App1}, Eq. (\ref{Eq7}) is equivalent to Eq. (\ref{Eq6}), Also, Eq. (\ref{Eq6}) is an approximation of the asymptotically optimal ML estimator Eq. (\ref{Eq2}). Therefore, the proposed estimator Eq. (\ref{Eq9}) is an approximation of the original ML estimator Eq. (\ref{Eq2}). 

\textit{Remark 3 (Convexity of Estimator)}: Compared to the ML estimator Eq. (\ref{Eq2}), Eq. (\ref{Eq9}) has no logarithms in the residuals after the above procedure. Nevertheless, it is still non-convex due to the non-convexity of the term $\beta_{i, j}^{2} /\left\|\boldsymbol{\theta_{j}}-\boldsymbol{\phi_{i}}\right\|^{2}$. Next, we will show how to develop a convex estimator based on the derived non-convex estimator Eq. (\ref{Eq9}). 

\textit{2) Establishment of Convex Estimator}: For simplicity, we define $k_{i}=\left\|\boldsymbol{\phi_{i}}\right\|^{2}$. According to Chebyshev norm \cite{boyd2004convex}, a closer examination of $\left\|\boldsymbol{\theta_{j}}-\bm{\phi_{i}}\right\|^{2}$ in estimator Eq. (\ref{Eq9}) reveals that it can be rewritten as
\begin{equation}
\left\|\boldsymbol{\theta_{j}}-\boldsymbol{\phi_{i}}\right\|^{2}=\boldsymbol{\theta_{j}}^{T} \boldsymbol{\theta_{j}}-2 \boldsymbol{\phi_{i}}^{T} \boldsymbol{\theta_{j}}+k_{i}
\label{Eq12}
\end{equation}

By introducing an auxiliary variable $\boldsymbol{\mu}=\left[\mu_{1, j}, \ldots, \mu_{N, j}\right]^{T}\left(\boldsymbol{\mu} \in \mathbb{R}^{N}\right)$ and $\boldsymbol{X}\left(\boldsymbol{X} \in \mathbb{S}^{2}\right)$, we can express estimator Eq. (\ref{Eq12}) in terms of $\boldsymbol{\mu}$ and $\boldsymbol{X}$ as \cite{ouyang2010received}
\begin{align}
&\left(\hat{\boldsymbol{\theta}}_{j}, \hat{\boldsymbol{X}}, \hat{\boldsymbol{\mu}}\right)=\arg \min _{\theta_{j}, \boldsymbol{X}, \boldsymbol{\mu}} f(\boldsymbol{\mu}) \notag\\
\text { s.t. } &\operatorname{tr}(\boldsymbol{X})-2 \boldsymbol{\phi_{i}}^{T} \boldsymbol{\theta_{j}}+k_{i} \leq \beta_{i, j}{ }^{2} \mu_{i, j} \notag\\
&\operatorname{tr}(\boldsymbol{X})-2 \boldsymbol{\phi_{i}}^{T} \boldsymbol{\theta_{j}}+k_{i} \geq \beta_{i, j}{ }^{2} \mu_{i, j}{ }^{-1} \notag\\
&\boldsymbol{X}=\boldsymbol{\theta_{j}} \boldsymbol{\theta_{j}}^{T} 
\label{Eq13}
\end{align}

Appendix \ref{App2} provides a detailed proof of Eq. (\ref{Eq13}) derived from Eq. (\ref{Eq9}). In the above formulation, the constraints $\operatorname{tr}(\boldsymbol{X})-2 \boldsymbol{\phi_{i}}^{T} \boldsymbol{\theta_{j}}+k_{i} \leq \beta_{i, j}{ }^{2} \mu_{i, j} \notag$ are affine constraints, and $\operatorname{tr}(\boldsymbol{X})-2 \boldsymbol{\phi_{i}}^{T} \boldsymbol{\theta_{j}}+k_{i} \geq \beta_{i, j}{ }^{2} \mu_{i, j}{ }^{-1} \notag$ are convex constraints since $\operatorname{tr}(\boldsymbol{X})$ are linear in $\boldsymbol{X}$, and $-2 \boldsymbol{\phi_{i}}^{T} \boldsymbol{\theta_{j}}$ is linear $\boldsymbol{\theta}_j$, and $\mu_{i, j}{ }^{-1}$ is convex on $\mu_{i, j}>0$. However, the equality constraint $\boldsymbol{X}=\boldsymbol{\theta_{j}} \boldsymbol{\theta_{j}}^{T}$ is not affine. Therefore, Eq. (\ref{Eq13}) is still not convex. 

To develop a convex estimator, we relax the equality constraint $\boldsymbol{X}=\boldsymbol{\theta_{j}} \boldsymbol{\theta_{j}}^{T}$ to an inequality constraint $\boldsymbol{X} \succeq \boldsymbol{\theta_{j}} \boldsymbol{\theta_{j}}^{T}$ (well-known semi-definite relaxation). There, we can express Eq. (\ref{Eq13}) as a linear matrix inequality (LMI) \cite{boyd1994linear} by using a Schur complement \cite{boyd2004convex}.

Hence, the CV2X-based vehicle localization problem can be relaxed to the following standard SDP problem, i.e.,  
\begin{align}
&\left(\hat{\theta}_{j}, \hat{\boldsymbol{X}}, \hat{\boldsymbol{\mu}}\right)=\arg \min _{\theta_{j}, X, u} f(\boldsymbol{\mu})\notag\\
\text { s.t. } &\operatorname{tr}(\boldsymbol{X})-2 \boldsymbol{\phi_{i}}^{T} \boldsymbol{\theta_{j}}+k_{i} \leq \beta_{i, j}{ }^{2} \mu_{i, j} \notag\\
&\operatorname{tr}(\boldsymbol{X})-2 \boldsymbol{\phi_{i}}^{T} \boldsymbol{\theta_{j}}+k_{i} \geq \beta_{i, j}{ }^{2} \mu_{i, j}{ }^{-1} \notag\\
&{\left[\begin{array}{cc}
\boldsymbol{X} & \boldsymbol{\theta_{j}}\\
\boldsymbol{\theta_{j}^{T}} & 1
\end{array}\right] \succeq 0}
\label{Eq14}
\end{align}

Eq. (\ref{Eq14}) is convex and can be easily solved using a developed numerical toolbox \cite{sturm1999using}. Thus, we can guarantee that the solution to the problem is also the global minimum. 

Similarly, we can also express $\operatorname{tr}(\boldsymbol{X})-2 \boldsymbol{\phi_{i}}^{T} \boldsymbol{\theta_{j}}+k_{i} \geq \beta_{i, j}{ }^{2} \mu_{i, j}{ }^{-1} \notag$ as LMIs \cite{boyd1994linear}, i.e.,
\begin{align}
\left[\begin{array}{cc}
\operatorname{tr}(\boldsymbol{X})-2 \boldsymbol{\phi_{i}}^{T} \boldsymbol{\theta_{j}}+k_{i} & \beta_{i, j} \\
\beta_{i, j} & \mu_{i, j}
\end{array}\right] \succeq 0
\label{Eq15}
\end{align}

An excellent property of converting Eq.  (\ref{Eq14}) to Eq.  (\ref{Eq15}) is that $\mu_{i, j}^{-1}$ is circumvented in the expression, and the constraints become linear.

\textit{Remark 4 (Equivalence of Relaxation)}: The only difference between Eq. (\ref{Eq13}) and Eq. (\ref{Eq14}) is that we relax the equation constraint in Eq. (\ref{Eq13}) to an inequality constraint in Eq. (\ref{Eq14}). Thus, if the solution of Eq. (\ref{Eq14}) satisfies $\hat{\boldsymbol{X}}=\boldsymbol{\theta_{j}} \boldsymbol{\theta_{j}}^{T}$, we can conclude that $\boldsymbol{\theta_{j}}$ given by Eq. (\ref{Eq14}) is the global minimizer of Eq. (\ref{Eq13}), and also the global minimum of Eq. (\ref{Eq9}) (since Eq. (\ref{Eq13}) is equivalent to Eq. (\ref{Eq9})). If not, $\hat{\boldsymbol{\theta_{j}}}$ given by Eq. (\ref{Eq14}) is still feasible for Eq. (\ref{Eq9}), except $\boldsymbol{\theta_{j}}=\boldsymbol {\phi_{i}}$, since Eq. (\ref{Eq9}) is unconstrained with domain $\left\{\boldsymbol{\theta_{j}} \mid \boldsymbol{\theta_{j}} \neq \boldsymbol{\phi_{i}}\right\}$ (it means that the vehicle is out of the road because RSUs are deployed at the edge of the road). Furthermore, $f(\hat{\boldsymbol{\mu}})$ given by Eq. (\ref{Eq14}) provides a lower bound on the optimal value of Eq. (\ref{Eq9}), since we solve a relaxed problem on a larger set.

\subsection{Environment Parameter Correcting Module} \label{III-D}
\textit{1) Cooperative Roadside Units:} In the coarse positioning module, assuming that RSUs know all channel parameters may be impractical as it would require additional hardware. A common approach is to set a reference point (1 m here) to approximate the transmitted power $P_{0}$. However, $\beta_{i,j}$ remains uncertain due to the existence of $\gamma$, which is related to the surrounding environment. To address this issue, we propose a novel concept, called C-RSUs, where RSUs can communicate with each other in a dynamic C-V2X network while their location coordinates are known. As illustrated in Fig. \ref{fig3} (c), $RSU_1$, $RSU_2$, $RSU_3$, and $RSU_4$ can communicate with each other and are located within the same communication space environment as vehicles. By leveraging C-RSUs' capability for message sharing among themselves and capturing heterogeneity of traffic scenarios, the module can correct $\gamma$ using received signal power measured by one or more RSUs and reported back to others along with their location information.

\textit{2) Environmental Parameter Correction:} Assuming that the $i$th RSU can communicate with a group of surrounding RSUs, denoted as ``anchor nodes" for the sake of distinction. We define their coordinates using $\boldsymbol{\phi_{l}}=\left[b_{l, 1}, b_{l, 1}\right]^{T}\left(\boldsymbol{\phi_{l}} \in \mathbb{R}^{2}, l=1, \ldots, L\right)$ similar to the definition in section \ref{II-A}. It is important to note that $L$, which represents the number of anchor nodes, is independent from $N$. Therefore, it is possible for $L$ to be greater than ($L>N$), less than ($L<N$), or equal to ($L=N$) $N$. If channel parameters such as $P_0$ and $\gamma$ are known beforehand, then an unbiased maximum likelihood estimate of the distance between the $i$th RSU and each anchor node can be obtained as \cite{gholami2011positioning}
\begin{equation}
\hat{d}_{i, l}=d_0 10^{\frac{P_{i, l}-P_{0}}{10 \gamma_{i, l}}} e^{-\frac{10 \gamma_{i, l}}{\sigma_{i, l} \ln 10}}
\label{Eq16}
\end{equation}
where $d_{i, l}$ denotes the Euclidean distance between the $l$th anchor node and the $i$th RSU, and it can be computed by
\begin{equation}
d_{i, l}=\sqrt{\left(b_{l, 1}-b_{i, 1}\right)^{2}+\left(b_{l, 2}-b_{i, 2}\right)^{2}}
\label{Eq17}
\end{equation}

The Cramér-Rao lower-bound for the variance of any unbiased distance estimator based on RSS measurements can be obtained as \cite{gholami2011positioning}
\begin{equation}
\mathbb{E}\left(\hat{d}_{i, l}-\mathbb{E}\left(\hat{d}_{i , l}\right)\right)^2 \geq\left(\frac{\sigma_{i, l} \left\|\boldsymbol{\phi_l}-\boldsymbol{\phi_i}\right\| \ln 10}{10 \gamma_{i, l}}\right)^2
\label{Eq18}
\end{equation}

As previously stated, understanding all of the channel parameters can be challenging. An alternative approach is to make the assumption that $\sigma_{i,l}=\sigma_{dBm}$ for $l$=1,...,$L$, and if the interaction between $m_{i,l}$ and $\gamma_{i,l}$ is minimal, the biased estimate can be rewritten as
\begin{equation}
\hat{d_{i, l}}=d_0 10^{\frac{P_{i, l}-P_{0}}{10 \gamma_{i, l}}} 
\label{Eq19}
\end{equation}

Then, we can get the $\hat{\gamma_{i, l}}$ by using $L$ anchor nodes
\begin{equation}
\hat{\gamma_{i, l}}=\frac{P_{i, l}-P_{0}}{10 \log _{10} \hat{d_{i, l}}} (i=1, \ldots, N ; l=1, \ldots, L)
\label{Eq20}
\end{equation}

To minimize the impact of signal fluctuations, we calculate the average of $\hat{\gamma_{i,l}}$ from $L$ anchor nodes for a specific RSU ($i$th), which serves as our final estimate
\begin{equation}
\hat{\gamma}_{i}=\frac{1}{L} \sum_{l=1}^{L} \hat{\gamma_{i, l}}
\label{Eq21}
\end{equation}

\textit{Remark 5 (Rationality and Limitations)}: The noise $m_{i,j}$ in Eq. (\ref{Eq1}) is unmeasurable and cannot be disregarded. We make the assumption that the interaction between $m_{i,l}$ and $\gamma_{i,l}$ in Eq. (\ref{Eq19}) is small enough due to two reasons: (1) The received signal power of anchor nodes used for correction is instantaneous, resulting in small random noise between RSUs and anchor nodes; (2) $\sigma_{i,l}=\sigma_{dBm}$ indicates that transmitting power of each RSU is identical, leading to consistent noise distribution among them. Additionally, Eq. (\ref{Eq21}) further eliminates any random error. Strictly speaking, what we obtain from Eq. (\ref{Eq21}) is only an approximate solution to Eq. (\ref{Eq16}). Nevertheless, this approach of dynamically correcting environmental parameters more reasonable than using a fixed empirical value.

\subsection{Vehicle Trajectory Filtering Module} \label{III-E}
Based on the work of \cite{jo2016tracking}, we design a vehicle trajectory filtering module that incorporates prior traffic information, specifically vehicle motion features, to enhance lane-level position accuracy, as illustrated in Fig \ref{fig3} (e).

\textit{1) Vehicle Motion Model:} Jo \textit{et al.} have classified the vehicle motion model into four groups based on longitudinal and lateral motion features for regular road driving \cite{jo2016tracking}. For this research, we choose the constant velocity lane changing (CVLC) model as the vehicle motion model. The CVLC model assumes no acceleration or braking for the vehicle motion. The vehicle state space of the CVLC model can be defined as 
\cite{jo2016tracking}
\begin{equation}
\vec{\theta}=\left[\begin{array}{llll}x & v_{x} & y & v_{y}\end{array}\right]^{T}
\label{Eq22}
\end{equation}
where $x$ denotes lateral position, $y$ denotes longitudinal position, $v_{x}$ and $v_{y}$ correspond to the lateral speed and longitudinal speed of the vehicle in the running direction, respectively. Additionally, $\vec{\theta}$ denotes the ideal state of the vehicle.

Assume the state transition equation as follows:
\begin{equation}
\vec{\theta}(t+\Delta t)=\Delta f(t)+\vec{\theta}(t)
\label{Eq23}
\end{equation}
where $\Delta t$ represents the frequency of detection, and $\Delta f(t)$ in the state transition equation is defined by Eq. (\ref{Eq24})
\begin{equation}
\Delta f(t)=\left[\begin{array}{llll}
1 & 0 & \Delta t & 0 \\
0 & 1 & 0 & \Delta t \\
0 & 0 & 1 & 0 \\
0 & 0 & 0 & 1
\end{array}\right]
\label{Eq24}
\end{equation}

\textit{2) Filter Considering Uncertainty:} Vehicle trajectory prediction should account for uncertainty arising from measurement noise and processing noise. In this study, the vehicle trajectory filtering problem is considered nonlinear and be described as
\begin{equation}
\vec{\theta}(t+\Delta t)=f(\vec{\theta}(t), t)+Q(t)
\label{Eq25}
\end{equation}
\begin{equation}
\vec{z}(t)=h(\vec{\theta}(t), t)+R(t)
\label{Eq26}
\end{equation}
where $f$ is the motion function, $Q$ is the system noise (defined as the Gaussian noise), $\vec{z}$ is the observation parameters, $h$ is the observation function, and $R$ is the observation noise.

Several techniques have been devised to address uncertainty, including KF, EKF, and UKF. Jondhale \textit{et al.} \cite{jondhale2018kalman} demonstrated that the tracking performance of UKF is better than KF/EKF in wireless sensing networks. Here, we utilize the UKF to handle uncertainty in the vehicle trajectory filter. The relationship between the vehicle's estimated location obtained from Eq. (\ref{Eq14}) and  its location after UKF can be expressed as
\begin{equation}
[\hat{x}, \hat{y}]^T=g_{ukf}(\hat{\boldsymbol{\theta}})
\label{Eq27}
\end{equation}
where $g_{ukf}$ refers to the UKF funtion. The resulting output of the UKF, denoted as $[\hat{x}, \hat{y}]^T$, is considered as the final vehicle's position coordinate.

\section{Simulation Experiment} \label{IV}
\subsection{Simulation Setup}
\begin{table}[]
\begin{center}
\caption{Parameter Setting in the Simulation Experiments.}
\begin{tabular}{c|c|c}
\hline
\textbf{Symbol} & \textbf{Parameters}          & \textbf{Value} \\ \hline
$d_0$               & Reference distance           & 1m             \\
$v_x$               & Vehicle speed                & 25km/h         \\
$N$               & Number of hearable RSUs      & 3              \\
$\sigma_{d B}^2$   & Shadowing standard deviation & 2dBm           \\
$\Delta t$               & Detection frequency          & 0.1s           \\
$d r_1$               & Deployment spacing           & 60m            \\
$d r_2$               & Distance                     & 1m             \\
$d r_3$              & Width of road                & 14m            \\
$L$              & The number of anchor nodes   & 4              \\ \hline
\end{tabular}
\label{Tab2}
\end{center}
NOTE: The optimal parameters for each method can be found in demo code.
\end{table}
We conducted numerical simulations using MATLAB R2019a to compare our CV2X-LOCA method, labeled as ``Ours", with other state-of-the-art localization methods (see Table \ref{Tab1}). We also compared the performance of ML-True, which provides the optimal solution for signal-based localization problems in a static sensor network. It is worth noting that this method uses the actual vehicle location as the initial point of iteration. The simulation was carried out on a two-way four-lane road segment, 2000m in length and with a standard lane width of 3.5m (as recommended in the Guide to Geometric Design of Highways \cite{gibreel1999state,li2018rse}). To ensure better angle diversity of measurements and lower lateral dilution of precision \cite{del2019network}, RSUs were deployed on both sides of the road with identical deployment parameters ($dr_1$, $dr_2$, and $dr_3$). We assumed that all RSUs had the same mounting height/orientation and used directional antennas with conical-like radio patterns. OBUs were elevated at a height of 1 m above ground level but below where RSUs would be deployed \cite{klapez2020application}. For simplicity, we assumed constant vehicle speed for both freeway and urban scenarios. Although RSUs can handle data in both directions, we only present experimental results for one direction for convenience purposes. Unless otherwise specified, Table \ref{Tab2} shows the settings used in our simulations.

Errors in positioning can occur due to inaccuracies in measuring motion vectors and signal noise. To ensure a consistent standard for comparing method performance, we have chosen average localization error (ALE), root mean square error (RMSE), mean absolute error (MAE), and mean absolute percentage error (MAPE) as evaluation metrics. Our evaluation criteria differs from Li \textit{et al.} \cite{li2018rse} and Qin \textit{et al.}\cite{qin2017vehicles}, who only consider longitudinal errors perpendicular to the road, by taking into account 2D coordinate errors that include both longitudinal and lateral errors, which is more realistic in real-world traffic scenarios. We conducted 100 simulation runs for each scenario and adjusted the parameters of each method for optimal performance, including the noise covariance matrix $Q$, measurement noise covariance matrix $R$, and covariance matrix $Z_0$ used in UKF.

\begin{align}
R &=\left[\begin{array}{cccc}
2.2 & 0 & 0 & 0 \\
0 & 1.2 & 0 & 0 \\
0 & 0 & 0.9 & 0 \\
0 & 0 & 0 & 0.5
\end{array}\right] \notag\\
Z_{0} &=\left[\begin{array}{cccc}
0.25 & 0 & 0 & 0 \\
0 & 0.4 & 0 & 0 \\
0 & 0 & 0.2 & 0 \\
0 & 0 & 0 & 0.01
\end{array}\right], Q=I_{4 \times 4}
\label{Eq28}
\end{align}

\begin{figure*}[!t]
\centering
\subfloat[]{\includegraphics[width=0.245\textwidth,height=1.45in]{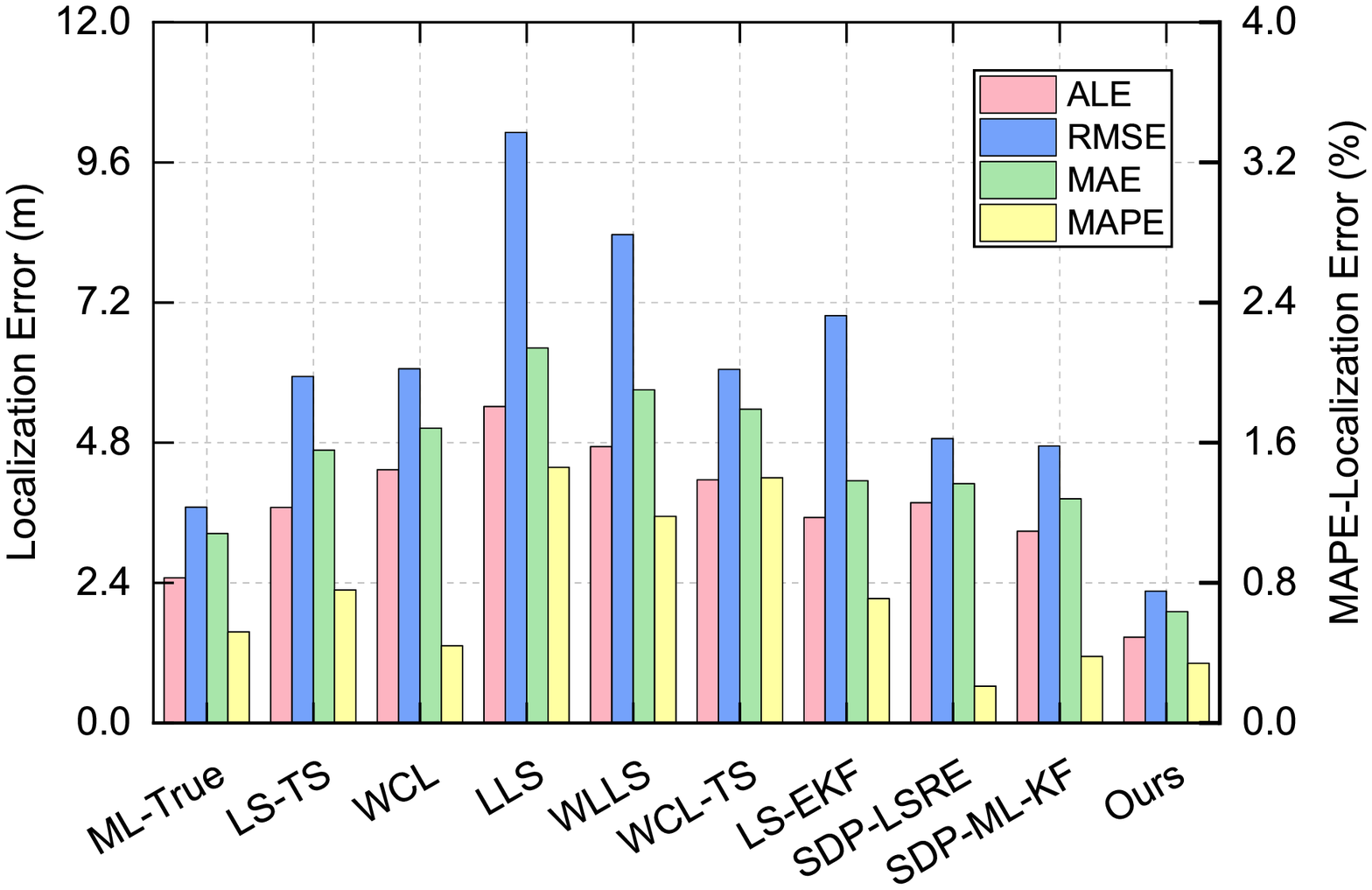}
\label{fig4_1}}
\subfloat[]{\includegraphics[width=0.245\textwidth,height=1.45in]{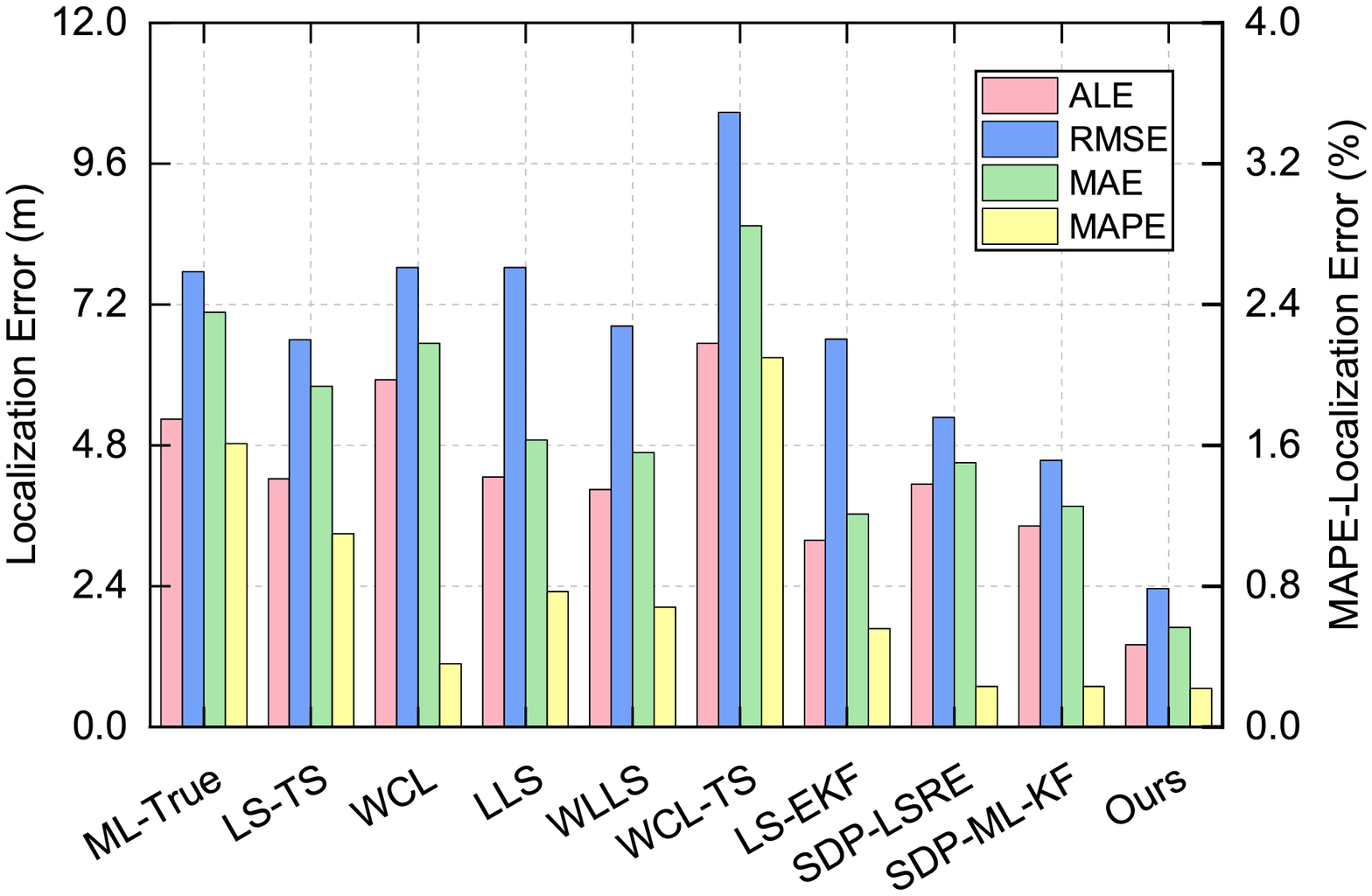}
\label{fig4_2}}
\subfloat[]{\includegraphics[width=0.245\textwidth,height=1.45in]{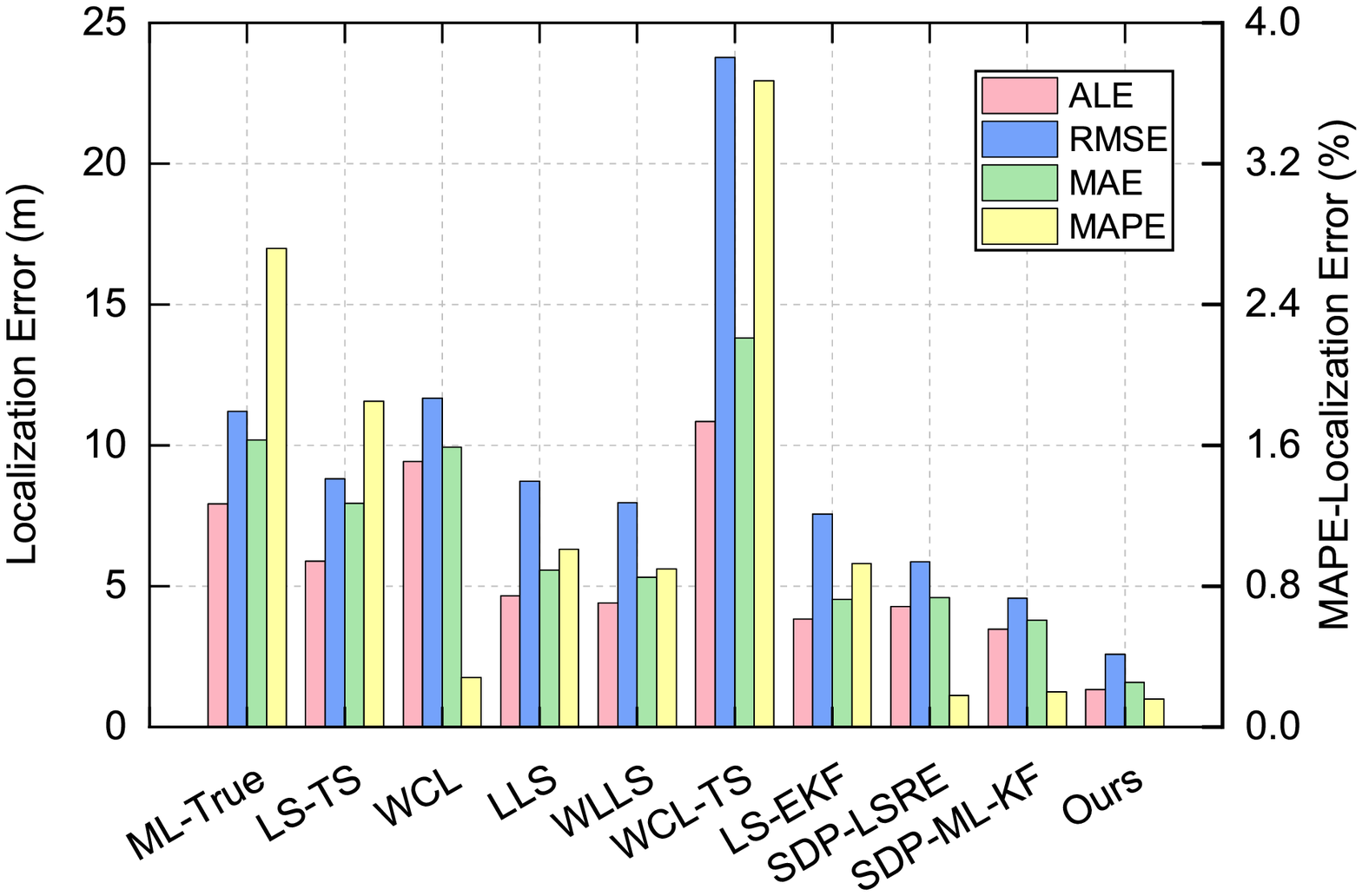}
\label{fig4_3}}
\subfloat[]{\includegraphics[width=0.245\textwidth,height=1.45in]{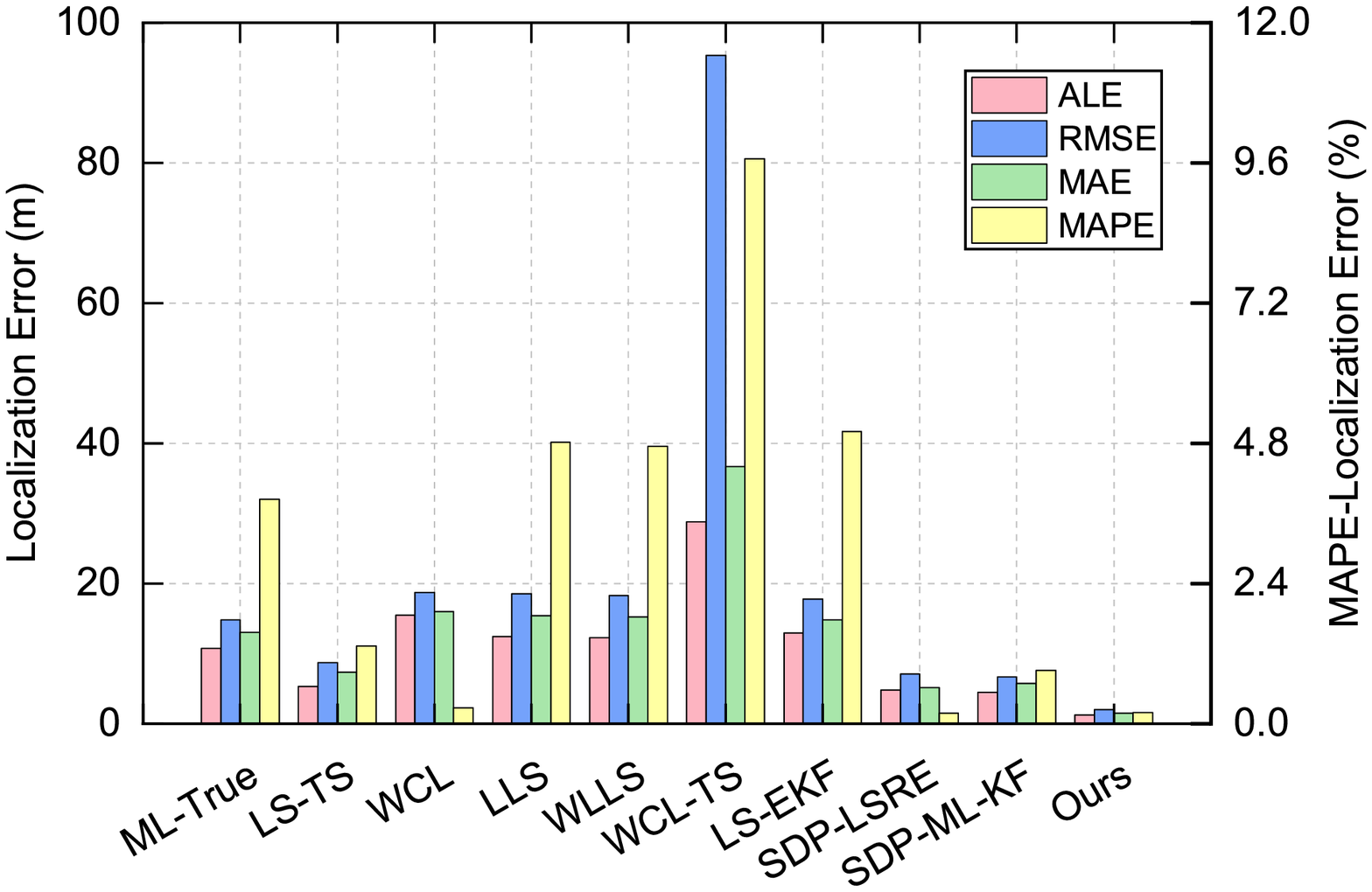}
\label{fig4_4}}

\subfloat[]{\includegraphics[width=0.245\textwidth,height=1.45in]{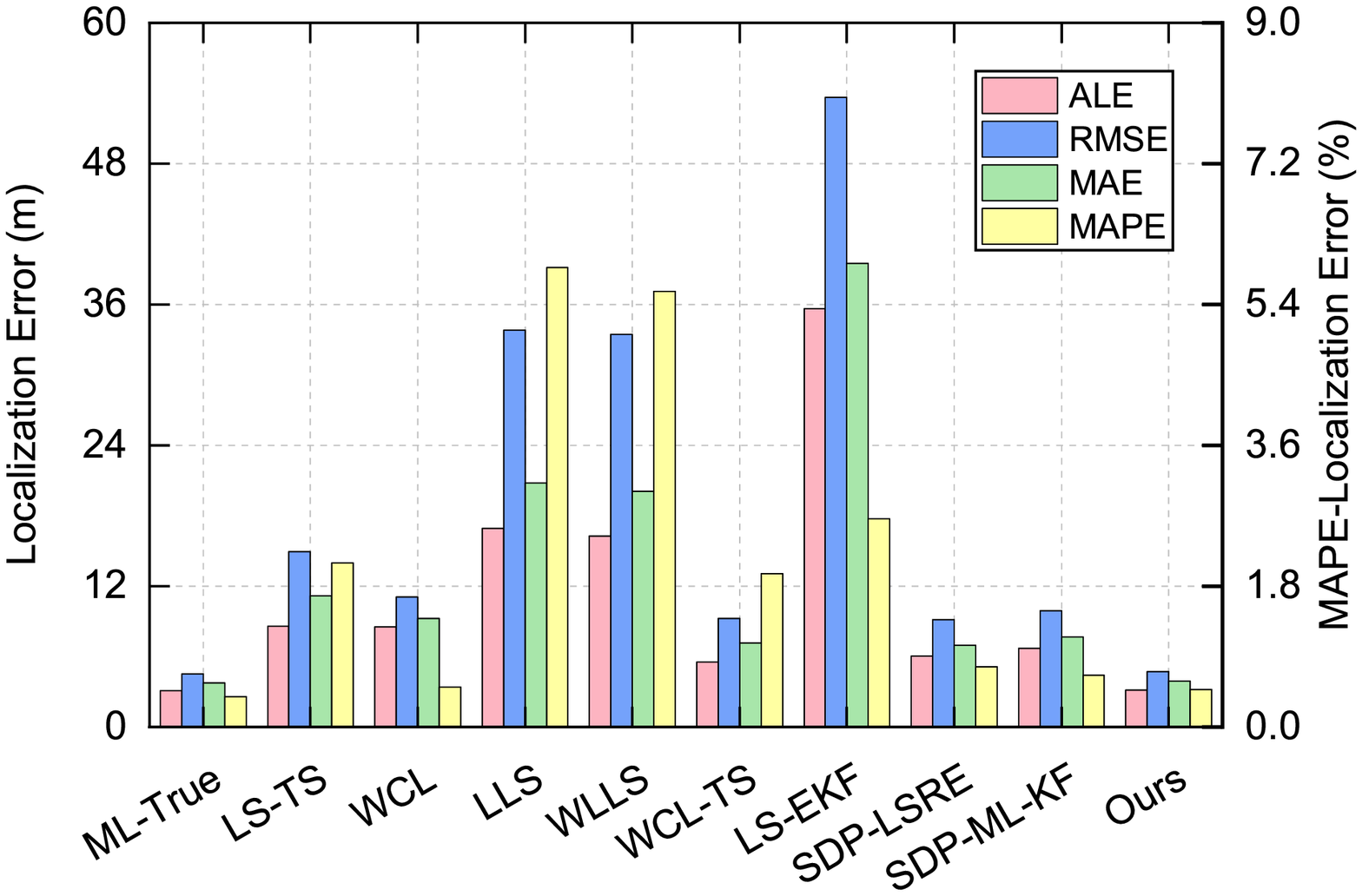}
\label{fig4_5}}
\subfloat[]{\includegraphics[width=0.245\textwidth,height=1.45in]{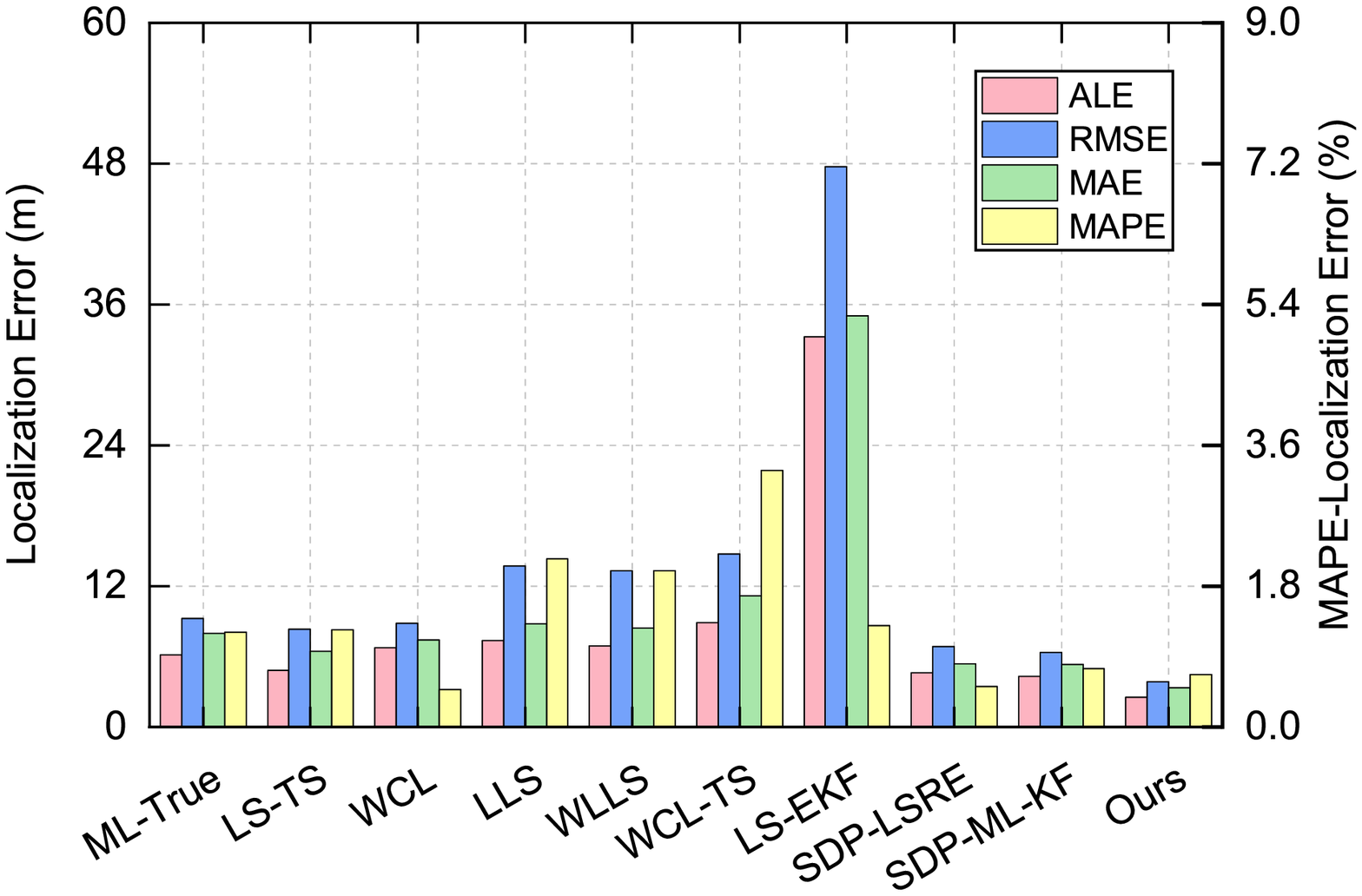}
\label{fig4_6}}
\subfloat[]{\includegraphics[width=0.245\textwidth,height=1.45in]{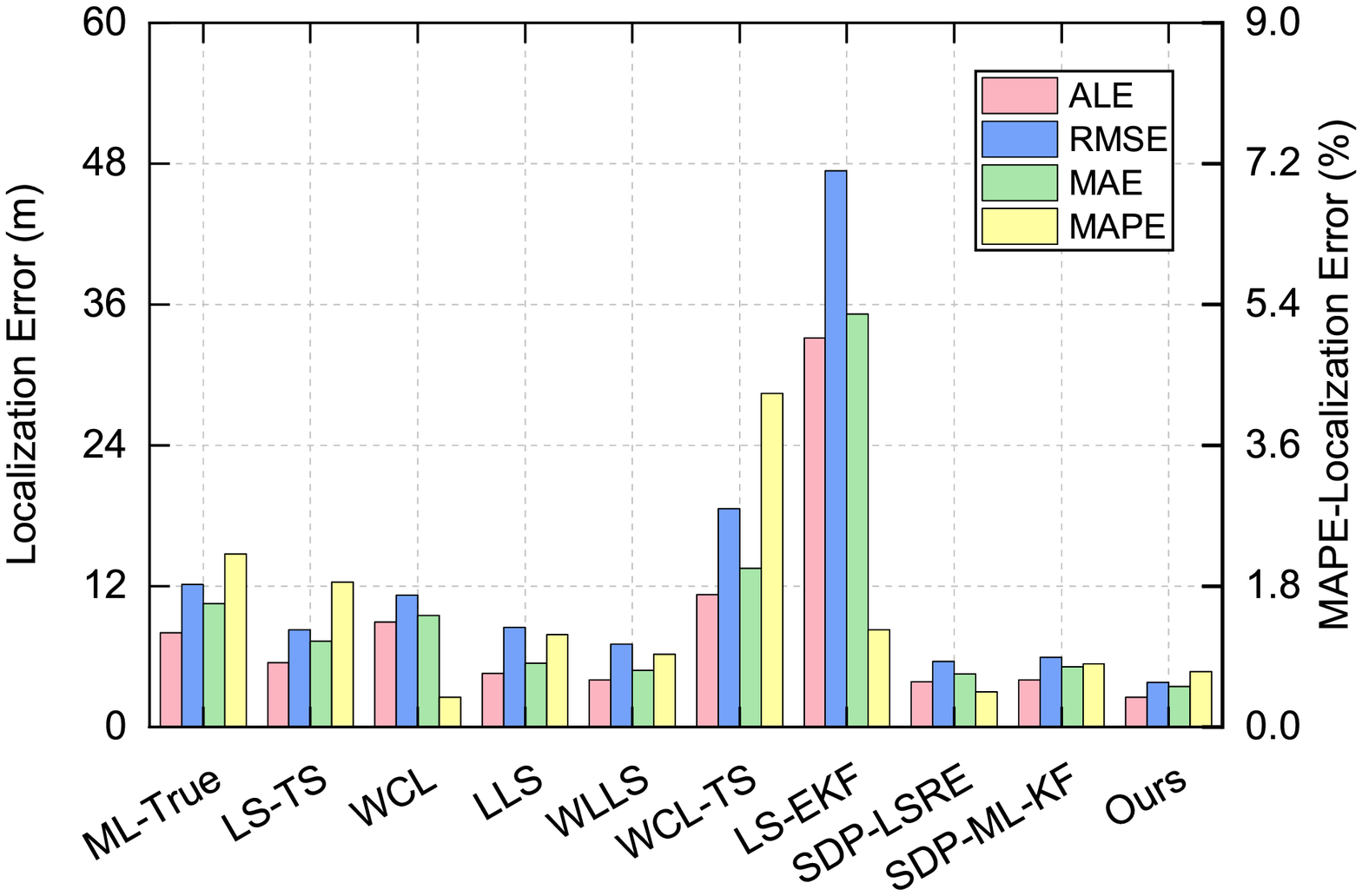}
\label{fig4_7}}
\subfloat[]{\includegraphics[width=0.245\textwidth,height=1.45in]{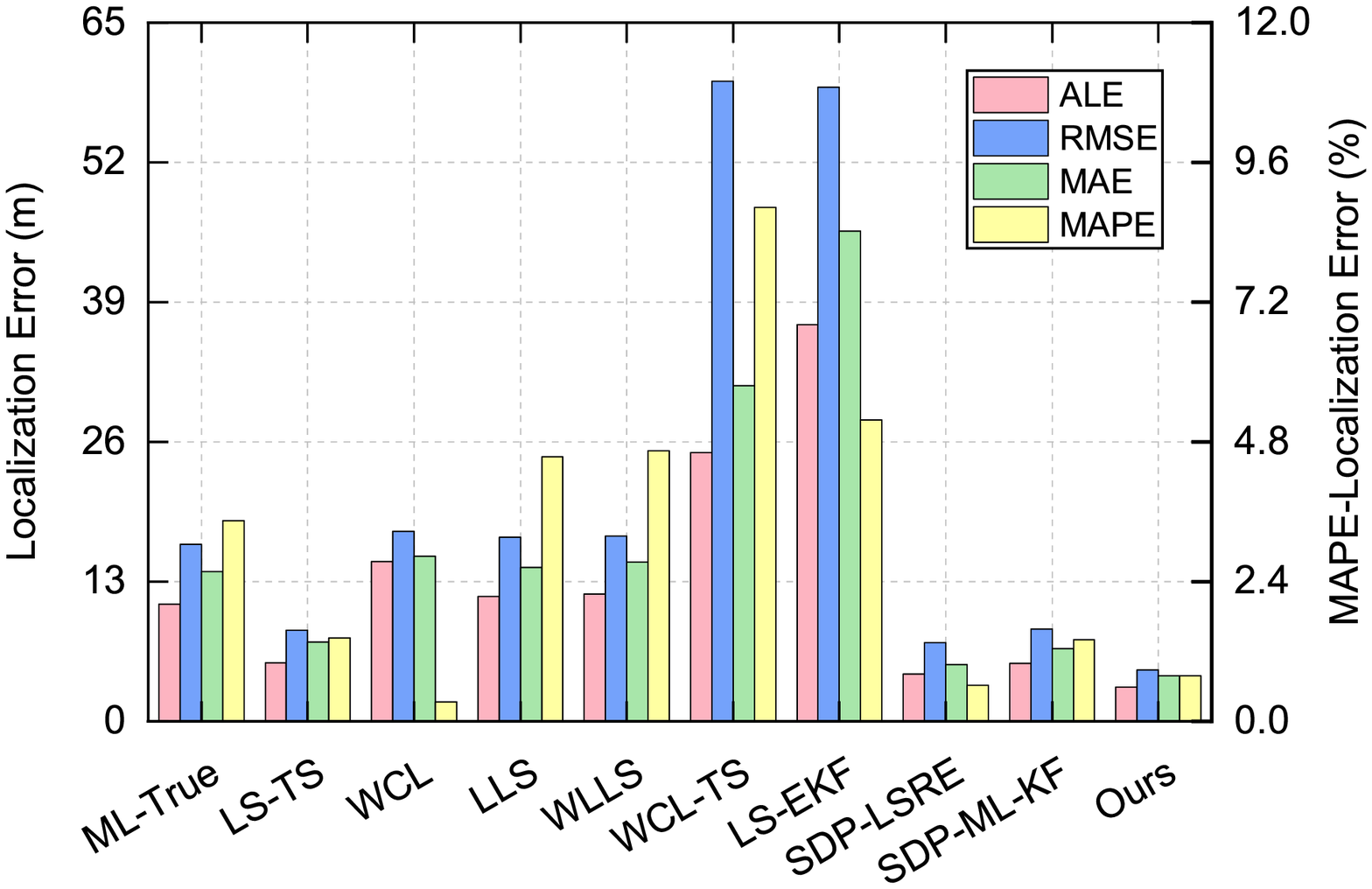}
\label{fig4_8}}
\caption{The localization performance of different methods in four road environments. (a) $\sim$ (d) denote road under an elevated road, road within urban forests, road beside skyscrapers, and tunnel, respectively (vehicle speeds of 25km/h). (e) $\sim$ (h) denote road under an elevated road, road within urban forests, road beside skyscrapers, and tunnel, respectively (vehicle speeds of 100km/h).}
\label{fig4}
\end{figure*}
\begin{table*} 
\begin{center}
\caption{Main Localization Performance Comparison.}
\scalebox{0.80}{
\begin{tabular}{c|cccc|cccc|cccc|cccc}
\hline
\textbf{Model}    & \textbf{ALE}  & \textbf{RMSE} & \textbf{MAE}  & \textbf{MAPE} & \textbf{ALE}  & \textbf{RMSE} & \textbf{MAE}  & \textbf{MAPE} & \textbf{ALE}  & \textbf{RMSE} & \textbf{MAE}  & \textbf{MAPE} & \textbf{ALE}  & \textbf{RMSE} & \textbf{MAE}  & \textbf{MAPE} \\ \hline
                  & \multicolumn{4}{c}{ Road Environment (i)}                             & \multicolumn{4}{c}{Road Environment (ii)}                             & \multicolumn{4}{c}{Road Environment (iii)}                             & \multicolumn{4}{c}{Road Environment (iv)}                             \\ \hline
\multicolumn{17}{c}{25km/h}                                                                                                                                                                                                                                                         \\ \hline
ML-True           & 2.48          & 3.7           & 3.25          & 0.52          & 5.25          & 7.76          & 7.07          & 1.61          & 7.93          & 11.21         & 10.19         & 2.72          & 10.78         & 14.83         & 13.04         & 3.84          \\
LS-TS \cite{jianyong2014rssi}            & 3.69          & 5.93          & 4.67          & 0.76          & 4.23          & 6.6         & 5.8         & 1.1           & 5.89          & 8.81          & 7.94          & 1.85          & 5.35          & 8.75          & 7.34          & 1.33          \\
WCL \cite{chen2016vehicle}              & 4.33          & 6.07          & 5.05          & 0.44          & 5.92          & 7.83         & 6.54          & 0.36          & 9.42          & 11.68         & 9.93          & 0.28          & 15.54         & 18.77         & 16.04         & 0.27          \\
LLS \cite{saeed2018localization}              & 5.42          & 10.11         & 6.42          & 1.46          & 4.26         & 7.83         & 4.89          & 0.77          & 4.65          & 8.74          & 5.57          & 1.01          & 12.47         & 18.58         & 15.39         & 4.82          \\
WLLS \cite{ma2019efficient}             & 4.73          & 8.37          & 5.71          & 1.18          & 4.04         & 6.84         & 4.68         & 0.68          & 4.4           & 7.98          & 5.31          & 0.9           & 12.31         & 18.33         & 15.24         & 4.75          \\
WCL-TS \cite{magowe2019closed}           & 4.16          & 6.06          & 5.38          & 1.4           & 6.54          & 10.47          & 8.54          & 2.1          & 10.84         & 23.77         & 13.82         & 3.67          & 28.81         & 95.31         & 36.66         & 9.67          \\
LS-EKF \cite{page2019enhanced}           & 3.52          & 6.97          & 4.15          & 0.71          & 3.18         & 6.61         & 3.63         & 0.56          & 3.84          & 7.57          & 4.54          & 0.93          & 12.99         & 17.82         & 14.81         & 5             \\
SDP-LSRE \cite{wang2018cooperative}         & 3.77          & 4.87          & 4.1           & 0.21          & 4.14          & 5.28         & 4.5          & 0.23          & 4.28          & 5.88          & 4.59          & 0.18          & 4.84          & 7.12          & 5.16          & 0.18          \\
GRNN-UKF \cite{jondhale2018kalman}         & 2.36          & 3.5           & 3.08          & 0.53          & 2.66         & 4.23         & 3.44         & 0.83          & 95.84         & 115.85        & 104.25        & 38.05         & 2756.53       & 3584.55       & 3001.42       & 1092.89       \\
CF-LS-UKF \cite{huang2020multi}        & 2.26          & 3.33          & 2.91          & 0.46          & 6.23         & 7.96         & 7.34         & 2.44          & 167.72        & 175.94        & 170.61        & 67.13         & 4255.55       & 4683.48       & 4344.25       & 1691.75       \\
SDP-ML-KF \cite{zhang2021localization}        & 3.28          & 4.74          & 3.84          & 0.38          & 3.43          & 4.55          & 3.76          & 0.23          & 3.49          & 4.57          & 3.79          & 0.2           & 4.47          & 6.68          & 5.78          & 0.92          \\
\textbf{CV2X-LOCA} & \textbf{1.47} & \textbf{2.26} & \textbf{1.91} & \textbf{0.34} & \textbf{1.4} & \textbf{2.36} & \textbf{1.69} & \textbf{0.22} & \textbf{1.33} & \textbf{2.58} & \textbf{1.59} & \textbf{0.16} & \textbf{1.28} & \textbf{2.06} & \textbf{1.54} & \textbf{0.19} \\ \hline
\multicolumn{17}{c}{100km/h}                                                                                                                                                                                                                                                       \\ \hline
ML-True           & 3.12          & 4.55          & 3.78          & 0.39          & 6.13          & 9.25          & 7.96          & 1.21          & 8.05          & 12.16         & 10.52         & 2.21          & 10.88         & 16.46         & 13.93         & 3.45          \\
LS-TS  \cite{jianyong2014rssi}             & 8.58          & 14.95         & 11.17         & 2.1           & 4.82          & 8.34          & 6.46          & 1.24          & 5.47          & 8.31          & 7.31          & 1.85          & 5.44          & 8.48          & 7.41          & 1.43          \\
WCL  \cite{chen2016vehicle}              & 8.53          & 11.11         & 9.27          & 0.51          & 6.74          & 8.87          & 7.4           & 0.48          & 8.97          & 11.24         & 9.5           & 0.38          & 14.89         & 17.69         & 15.38         & 0.34          \\
LLS  \cite{saeed2018localization}              & 16.94         & 33.81         & 20.8          & 5.87          & 7.38          & 13.74         & 8.8           & 2.15          & 4.58          & 8.49          & 5.42          & 1.18          & 11.63         & 17.13         & 14.34         & 4.55          \\
WLLS \cite{ma2019efficient}             & 16.29         & 33.44         & 20.09         & 5.57          & 6.9           & 13.31         & 8.44          & 2             & 4.03          & 7.07          & 4.81          & 0.93          & 11.85         & 17.25         & 14.8          & 4.65          \\
WCL-TS \cite{magowe2019closed}            & 5.55          & 9.23          & 7.16          & 1.96          & 8.91          & 14.74         & 11.21         & 3.28          & 11.3          & 18.6          & 13.55         & 4.26          & 25.02         & 59.53         & 31.23         & 8.83          \\
LS-EKF \cite{page2019enhanced}           & 35.65         & 53.66         & 39.51         & 2.66          & 33.27         & 47.77         & 35.06         & 1.3           & 33.17         & 47.4          & 35.19         & 1.24          & 36.91         & 58.98         & 45.62         & 5.18          \\
SDP-LSRE  \cite{wang2018cooperative}        & 6.05          & 9.16          & 6.99          & 0.77          & 4.64          & 6.87          & 5.39          & 0.52          & 3.84          & 5.59          & 4.51          & 0.45          & 4.43          & 7.32          & 5.29          & 0.62          \\
GRNN-UKF \cite{jondhale2018kalman}         & 13.57         & 19.24         & 15.86         & 4.82          & 5.56          & 7.88          & 7.12          & 2             & 88.71         & 126.13        & 105.87        & 32.86         & 2015.86       & 3293.06       & 2397.23       & 766.26        \\
CF-LS-UKF \cite{huang2020multi}        & 13.46         & 19.07         & 15.59         & 4.73          & 10.27         & 14.9          & 13.01         & 3.97          & 270.2         & 324.72        & 289.12        & 108.38        & 5124.46       & 6462.46       & 5466.99       & 2070.19       \\
SDP-ML-KF \cite{zhang2021localization}        & 6.72          & 9.91          & 7.66          & 0.66          & 4.33          & 6.34          & 5.34          & 0.75          & 4             & 5.95          & 5.15          & 0.81          & 5.41          & 8.58          & 6.78          & 1.4           \\
\textbf{CV2X-LOCA} & \textbf{3.17} & \textbf{4.73} & \textbf{3.93} & \textbf{0.48} & \textbf{2.54} & \textbf{3.85} & \textbf{3.37} & \textbf{0.67} & \textbf{2.56} & \textbf{3.81} & \textbf{3.46} & \textbf{0.71} & \textbf{3.18} & \textbf{4.8}  & \textbf{4.23} & \textbf{0.78} \\ \hline
\end{tabular}}
\label{Tab3}
\end{center}
\end{table*}

\subsection{Performance Comparison}
To examine the influence of road environments on localization performance, we analyzed four different types of road environments: (i) under an elevated road (semi-open environment), (ii) within urban forests, (iii) beside skyscrapers, and (iv) tunnel. Vehicle speeds were set at 25km/h and 100km/h. The results are presented in detail in Fig. \ref{fig4} and Table \ref{Tab3}. 

We observed some intriguing phenomena:

(1) We can find that the positioning error increases for all the methods in the higher speed condition. It may be that the moving vector measurement errors are sensitive to the rise in the travel distance, resulting from a higher speed, as they are measured at fixed time intervals. Despite this, CV2X-LOCA outperforms other benchmark methods in both lower and higher speed conditions with an ALE of less than 3.5m in all four road environments. These results demonstrate the robustness of our method in complex urban environments and its ability to provide lane-level positioning accuracy across different traffic road environments.

(2) Compared to LS-based methods (e.g., LS-TS, LLS, and LS-EKF), SDP-based techniques such as SDP-LSRE, SDP-ML-KF, and CV2X-LOCA are more effective in reducing positioning errors in the four road environments. This finding is consistent with the experimental results of Wang \textit{et al.} \cite{wang2018cooperative} and Zhang \textit{et al.} \cite{zhang2021localization}. Additionally, WLLS method has a lower localization error compared to LLS method which confirms the findings of Ma \textit{et al.} \cite{ma2019efficient}. We also observed that TS-based methods require better iterative initial points for optimal performance. For instance, while WCL-TS performs well in road environment (i), it fails in road environments (ii) $\sim$ (iv) where there is extensive error in the initial iterative point.

(3) ML-True demonstrates the highest localization performance in road environment (i) due to its semi-open nature, where obstacles only affect signals directly over the road. Furthermore, methods that incorporate vehicle motion features such as LS-EKF, SDP-ML-KF and our approach achieve greater positioning accuracy than those without. This indicates that integrating vehicle motion features can effectively enhance positioning accuracy. However, from the Table \ref{Tab3}, GRNN-UKF and CF-LS-UKF which excel in static sensor networks fail completely in road environments (iii) and (iv). This could be attributed to over-fitting of the modified signal-distance model and signal packet loss. For simplicity purposes, we have excluded these two methods from Fig. \ref{fig4} and they will not be compared further in subsequent sections.

The simulation road environment (i) is used as an example to assess the robustness and sensitivity of CV2X-LOCA.

\subsection{Robustness Evaluation}
\begin{figure}[!t]
\centering
\includegraphics[width=2.8in,height=2.1in]{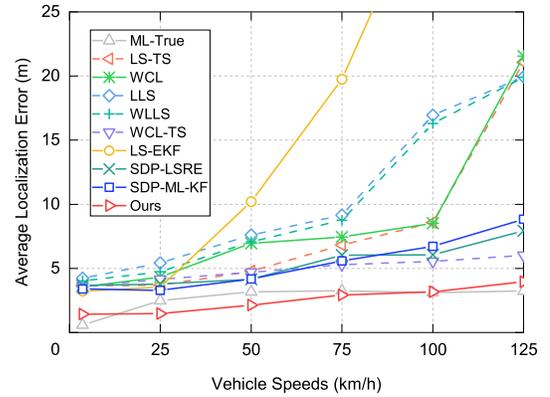}
\caption{The change of ALEs for different speeds (3.6km/h $\sim$125km/h)}
\label{fig5}
\end{figure}

\textit{1) Impacts of Vehicle Speed:} We tested six different speed values, ranging from 3.6 km/h to 125 km/h, to represent various driving conditions such as congestion, normal urban and rural driving, and highway driving. The changes of ALE for different vehicle speeds are compared across different methods in Fig. \ref{fig5}. As expected based on previous research by Liu \textit{et al.} \cite{liu2022research}, positioning error gradually increases with increasing vehicle speed for all methods but at varying rates. The LS-based methods (LS-TS and LS-EKF) show nearly exponential growth in errors with increasing speed, indicating poor adaptation to high-speed conditions. On the other hand, SDP-LSRE, SDP-ML-KF, and CV2X-LOCA has smaller errors and slower rates of error growth due to the advantages of the SDP framework in effectively overcoming noise and signal packet loss caused by speed changes. Among these methods, CV2X-LOCA demonstrates robustness against speed increase with the smallest positioning error and rate of growth. Notably, ML-True also maintains good robustness when using real vehicle location as an initial iteration point.

\begin{figure*}[!t]
\centering
\subfloat[]{\includegraphics[width=0.33\textwidth,height=1.7in]{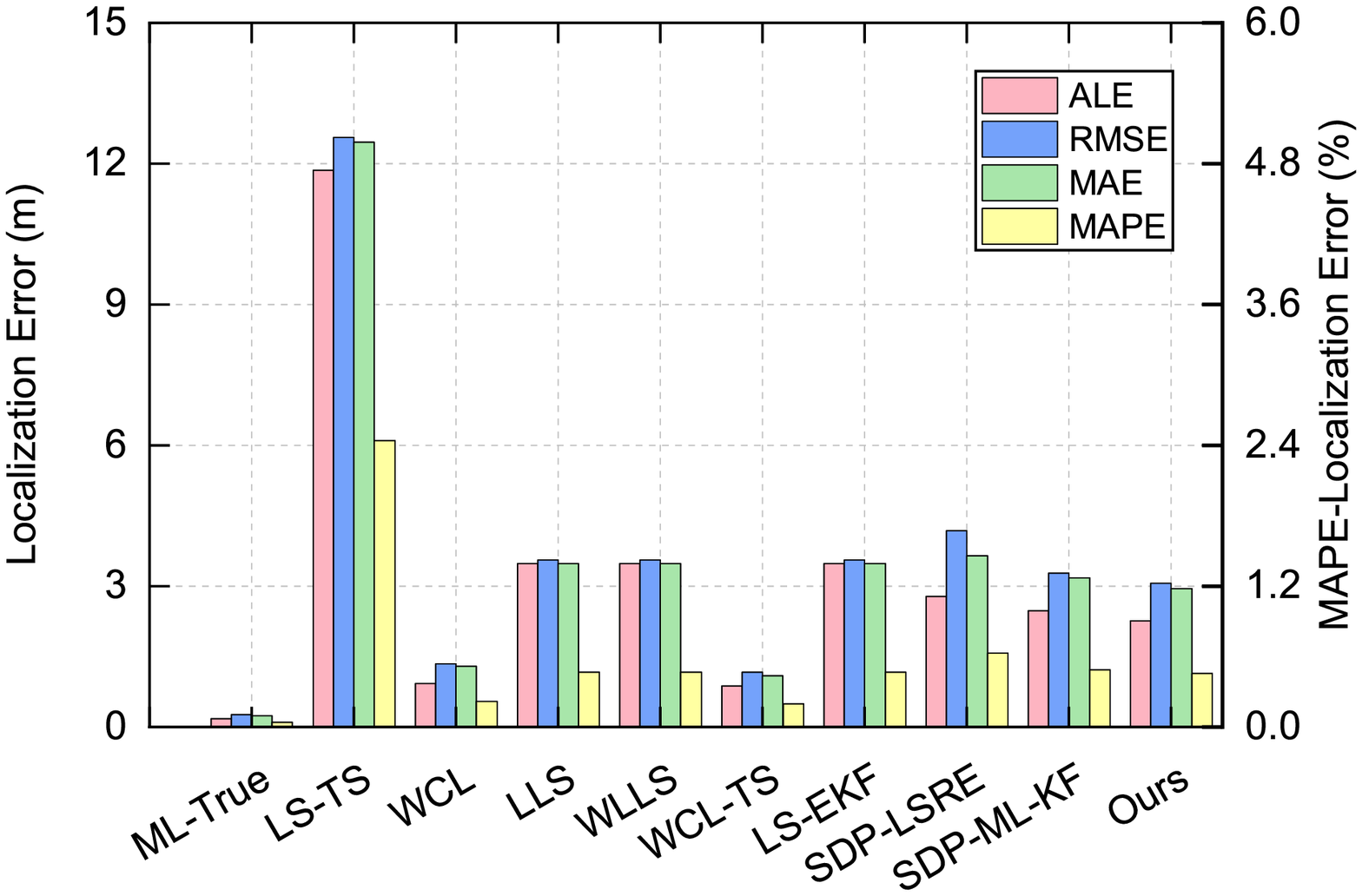}
\label{fig6_1}}
\subfloat[]{\includegraphics[width=0.33\textwidth,height=1.7in]{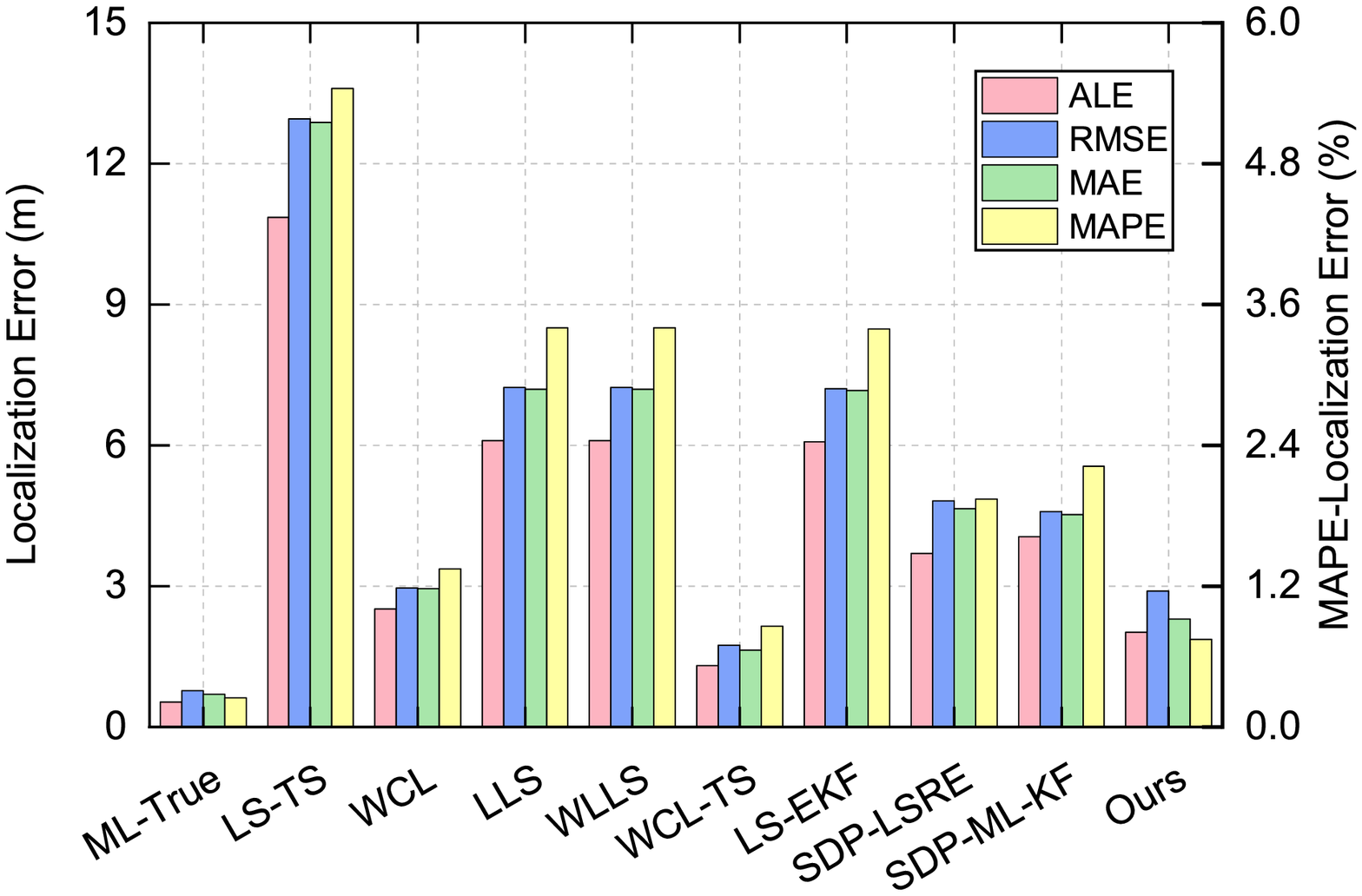}
\label{fig6_2}}
\subfloat[]{\includegraphics[width=0.33\textwidth,height=1.7in]{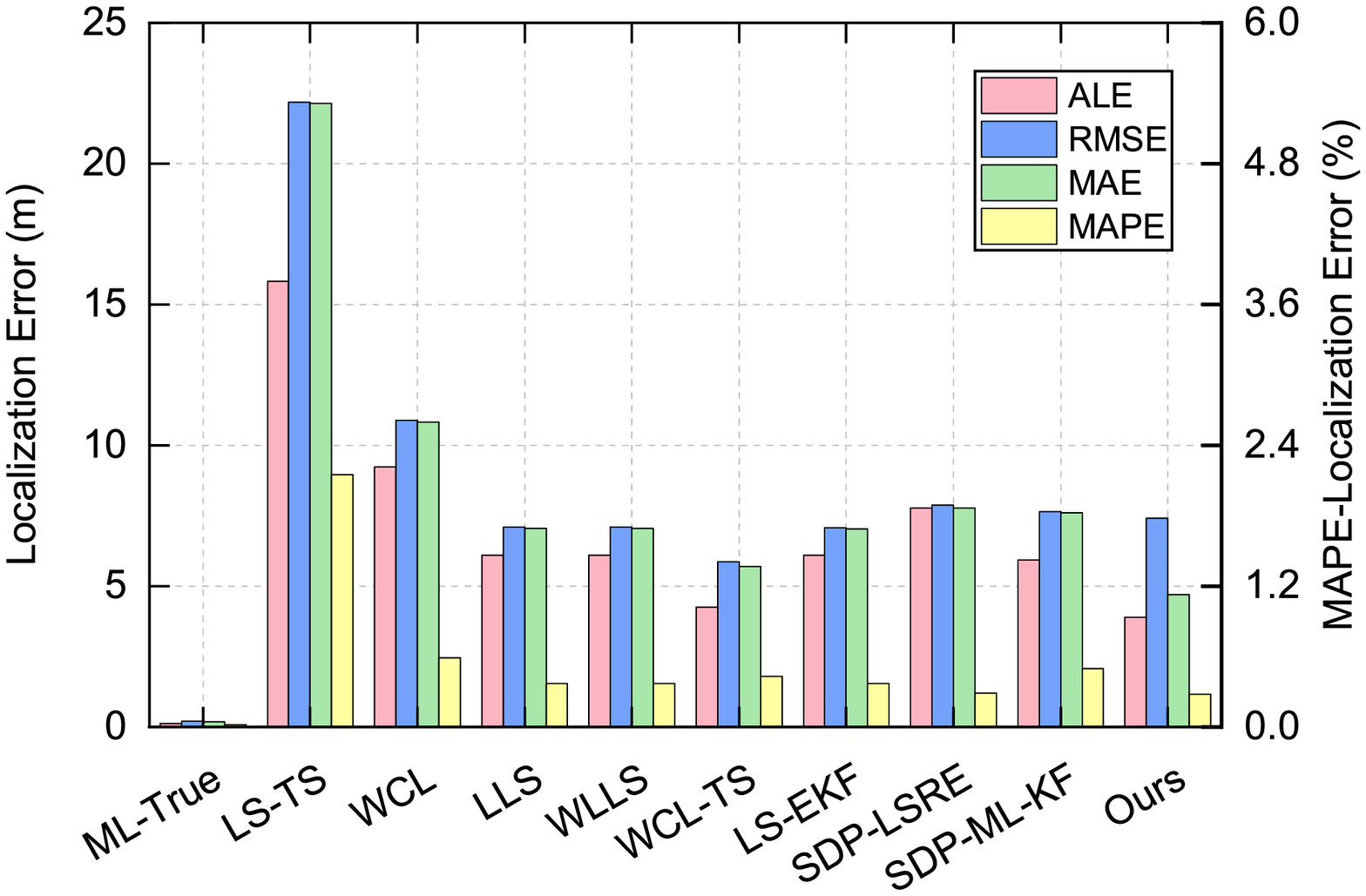}
\label{fig6_3}}
\vfill
\caption{The localization performance of different methods under three different geometric layouts. (a) vehicles are close to the centroid of the triangle formed by the RSUs. (b) vehicles are close to one RSU while being far away from the other two. (c) vehicles are outside the convex hull formed by the RSUs.}
\label{fig6}
\end{figure*}

\textit{2) Impacts of Geometric Layout of RSUs:} The accuracy of positioning is significantly affected by the geometric arrangement between vehicles and RSUs, known as the geometric dilution of precision problem \cite{ouyang2010received}. This consideration is crucial because signal packet loss or RSU failure is common in real-world traffic scenarios. We fixed the locations of the RSUs ($N$=3) and selected three typical vehicle locations relative to them: Location (i), which is close to the centroid of the triangle formed by the RSUs; Location (ii), which is near one RSU but far from the other two; and Location (iii), which lies outside the convex hull formed by the RSUs. The performance of different methods for these various locations can be seen in Fig. \ref{fig6}. When a vehicle is at location (i), there are only minor differences in performance across methods, possibly due to less signal fluctuation and packet loss when moving near the triangle's centroid. However, when a vehicle is at location (ii), LS-based methods' localization performance deteriorates \cite{wang2018cooperative}. In contrast, WCL and WCL-LS remain stable at both locations since they use weighted position information. At location (iii), all methods show more prominent ALEs, with CV2X-LOCA having lower positioning errors than others. Overall, our proposed CV2X-LOCA framework performs better than other SDP-based methods and proves robust against various vehicle-RSU layouts.

\begin{figure*}[!t]
\centering
\subfloat[]{\includegraphics[width=0.32\textwidth,height=1.7in]{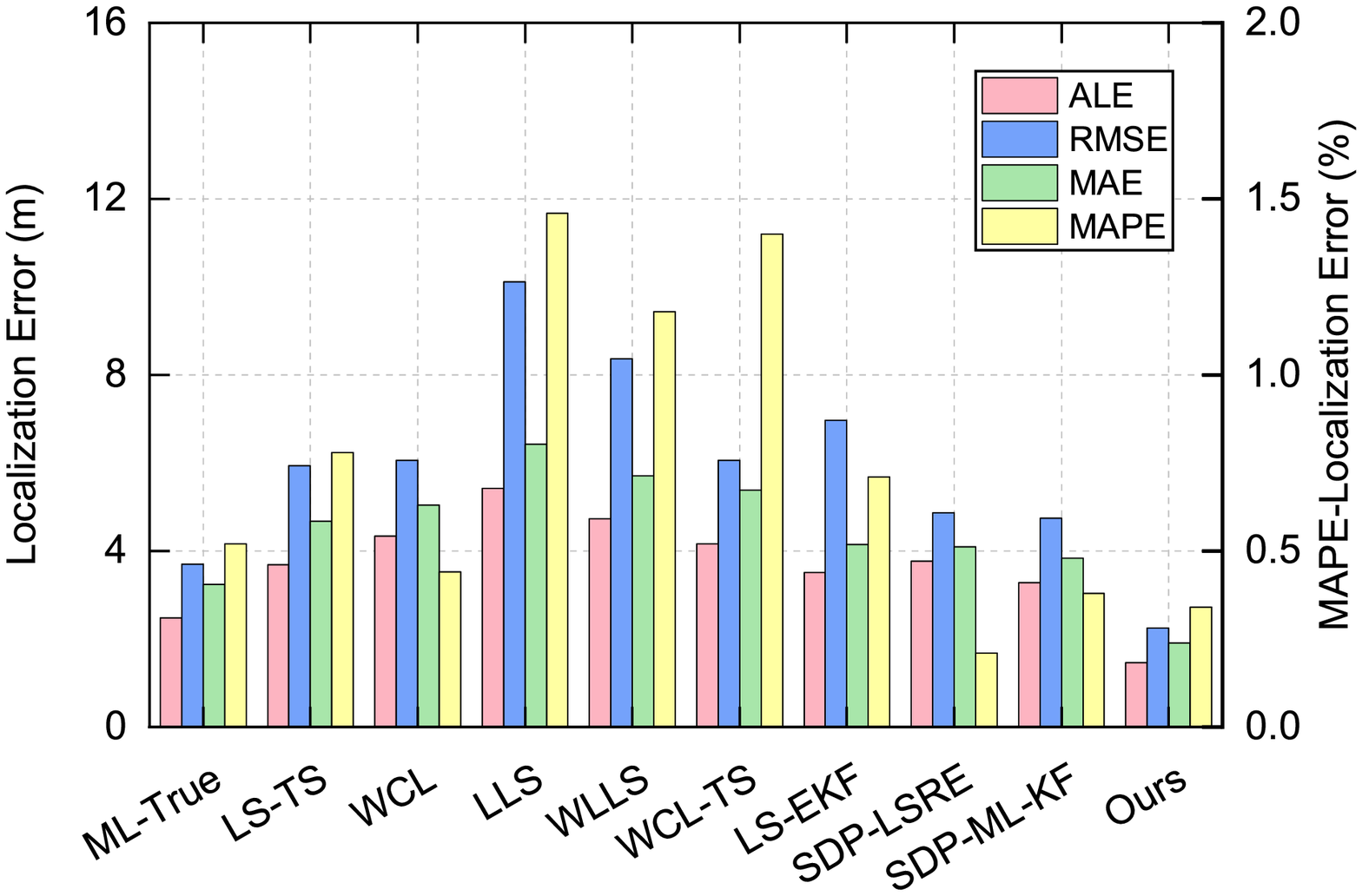}
\label{fig7_1}}
\subfloat[]{\includegraphics[width=0.32\textwidth,height=1.7in]{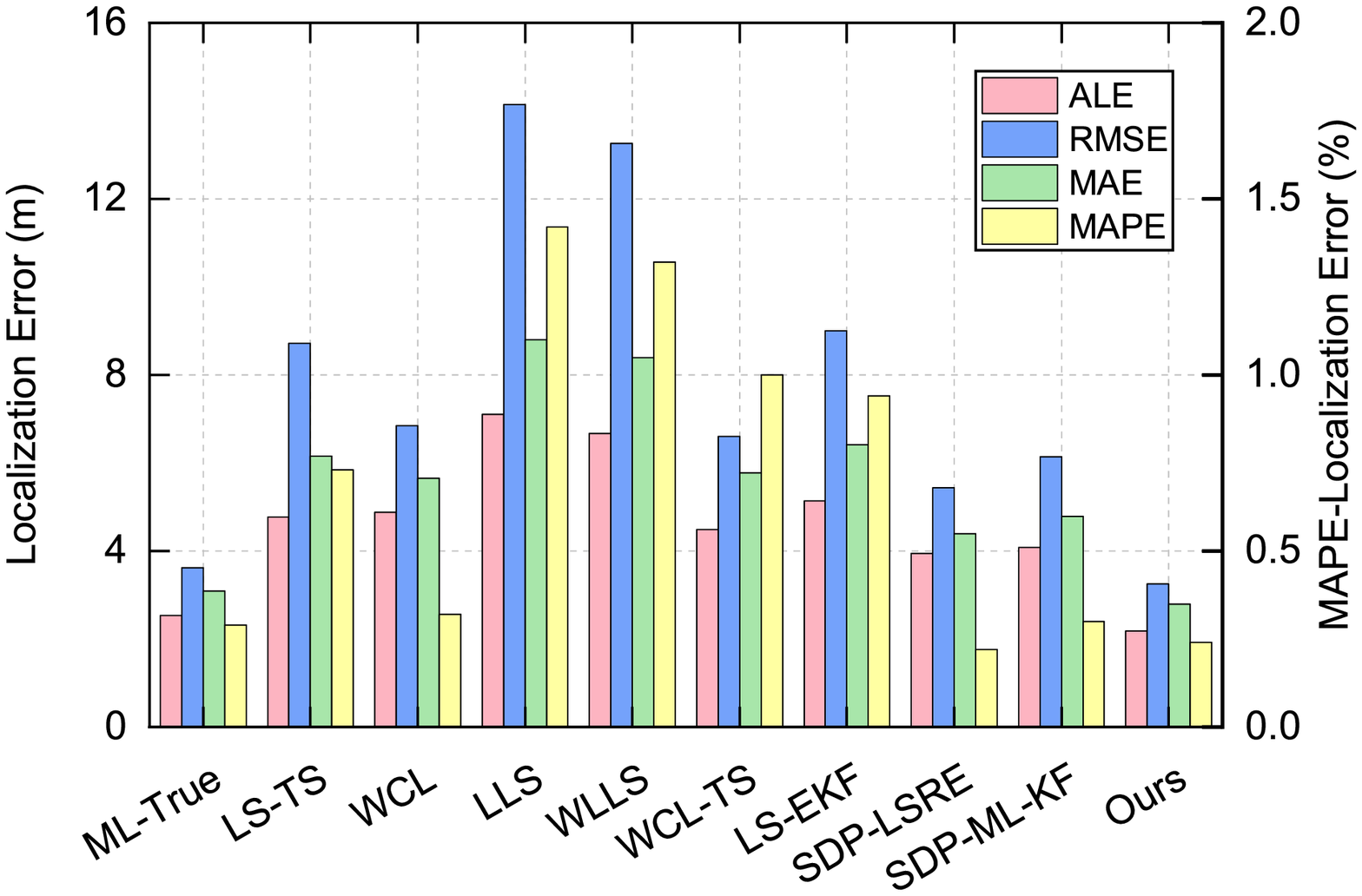}
\label{fig7_2}}
\subfloat[]{\includegraphics[width=0.32\textwidth,height=1.7in]{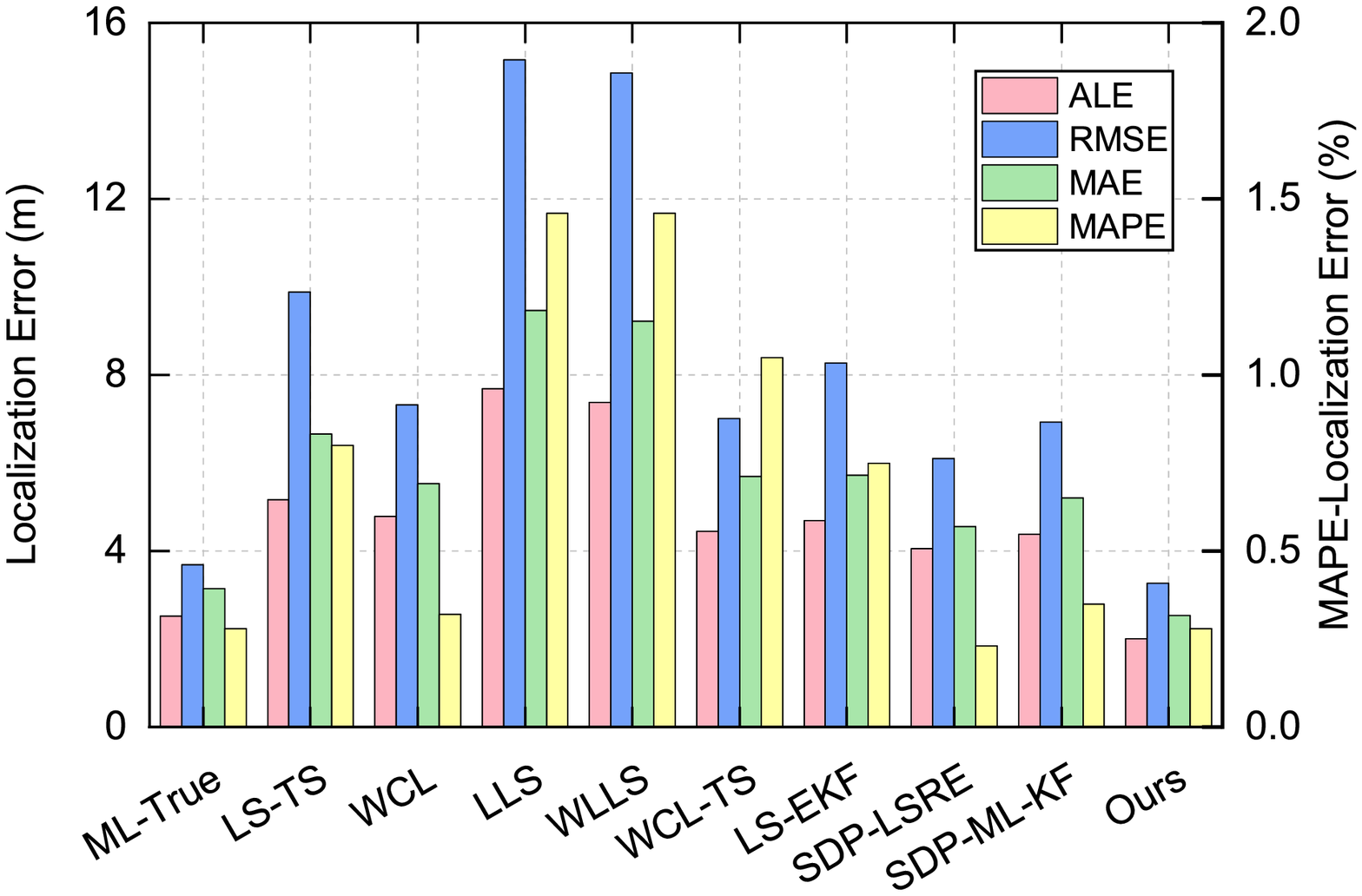}
\label{fig7_3}}
\vfill
\caption{The three types of trajectories and their corresponding localization errors of different localization methods. (a) vehicle moves without lane-changing. (b) vehicle switch from the right lane to the left lane. (c) vehicle switch from the left lane to the right lane.}
\label{fig7}
\end{figure*}

\textit{3) Impacts of Vehicle Lane-Changing:} Lane-changing is a common occurrence in real-world traffic situations, and it is crucial for the localization framework to maintain optimal performance during such maneuvers. Our study assumes three vehicle trajectories within a 2000m road segment: (i) the vehicle remains in the right lane without changing lanes; (ii) the vehicle changes lanes from the right lane to left lane between 400m and 580m; and (iii) the vehicle changes lanes from left to right at 400m up to 580m. The results of our study are presented in Fig. \ref{fig7}, which shows that all localization methods, except ML-True, experience a slight increase in positioning errors when vehicles change lanes. However, CV2X-LOCA outperforms other methods with positioning errors of only 1.47m, 2.19m, and 2.01m for trajectories (i) $\sim$ (iii) respectively. These findings demonstrate that our proposed method is more adaptable to complex driving behaviors and has promising application prospects.

\subsection{Sensitivity Evaluation}
\begin{figure*}[!t]
\centering
\subfloat[]{\includegraphics[width=0.31\textwidth,height=1.7in]{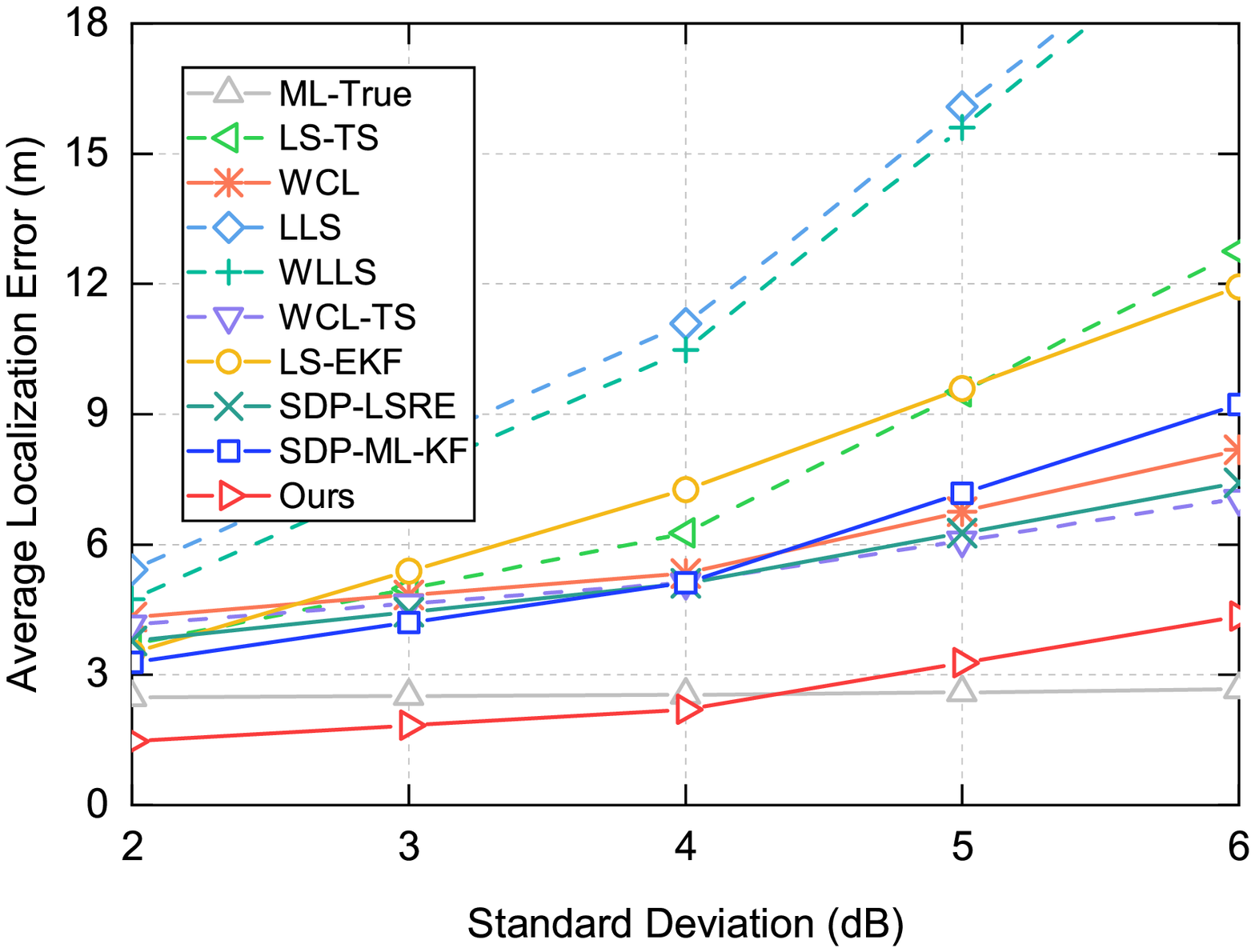}
\label{fig8_1}}
\quad
\subfloat[]{\includegraphics[width=0.31\textwidth,height=1.7in]{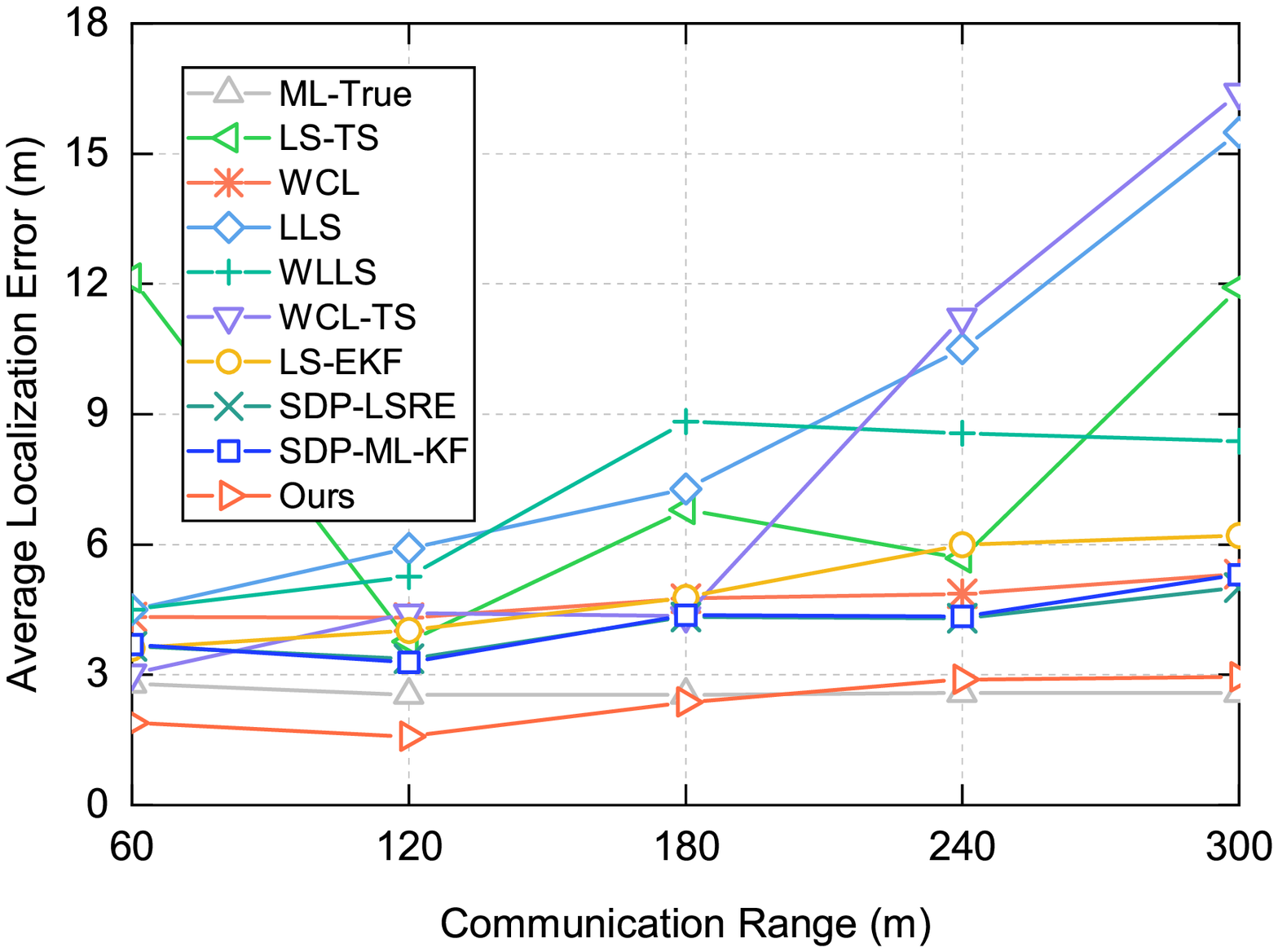}
\label{fig8_2}}
\quad
\subfloat[]{\includegraphics[width=0.31\textwidth,height=1.7in]{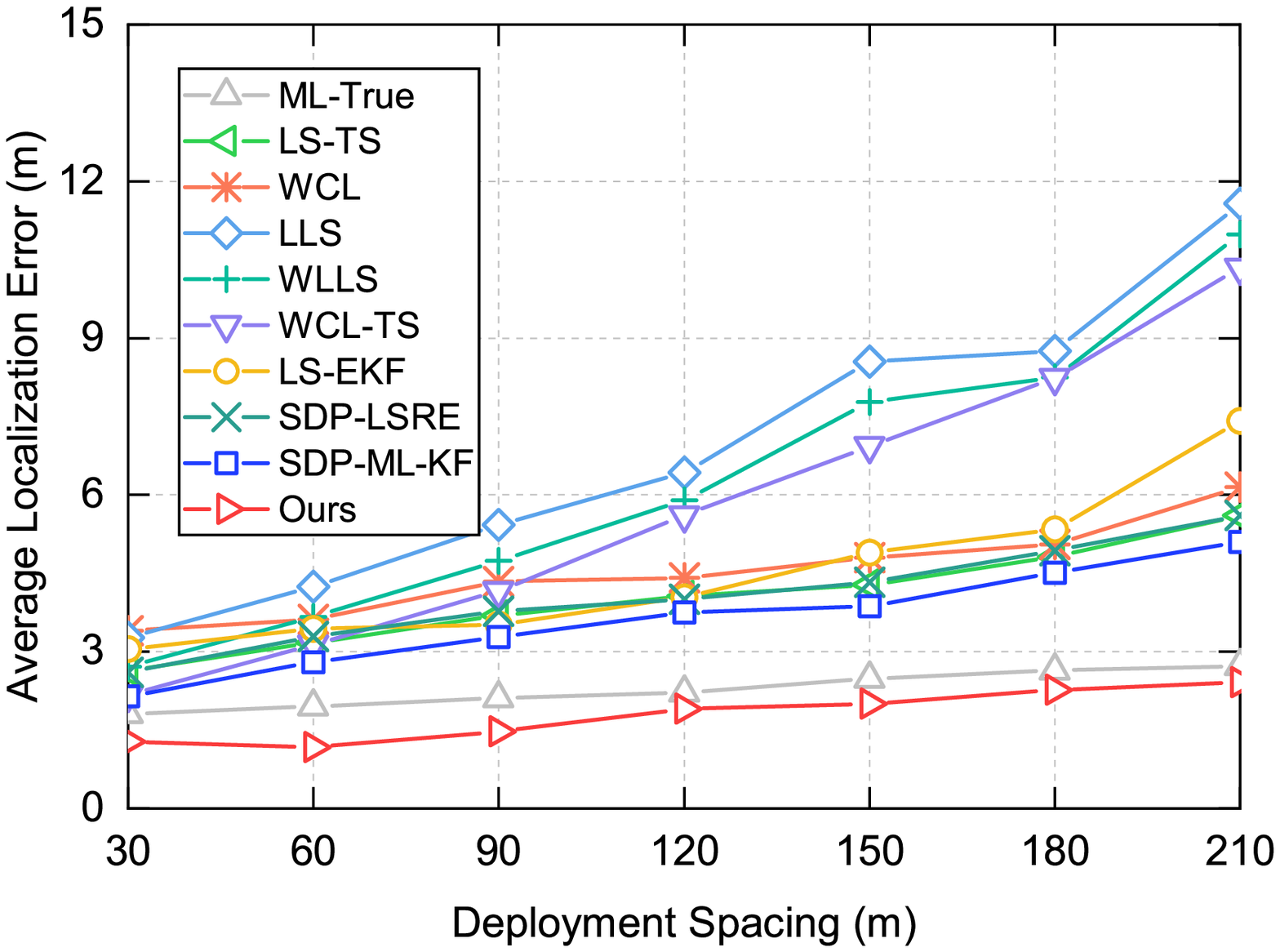}
\label{fig8_3}}
\caption{Sensitivity comparison of localization performance of different localization methods. (a) the change of ALEs under different standard deviations of noise (between 2 $\sim$ 6 dBm). (b) the change of ALEs under different communication ranges (between 60 $\sim$ 300m). (c) the change of ALEs under different deployment spacing (between 30 $\sim$ 210m).}
\label{fig8}
\end{figure*}

\textit{1) Impacts of Shadowing Standard Deviation:} In a dynamic C-V2X network, the communication signal value between RSUs and OBUs fluctuates due to vehicle shielding and other factors \cite{nguyen2022cellular}. The standard deviation of shading in natural traffic scenarios ranges from 2 to 6 \cite{linrssi}. Fig. \ref{fig8} (a) shows how different levels of noise affect the positioning performance. As expected, all localization methods experience an increase in ALE with higher standard deviations of shading. However, our method demonstrates slower performance degradation compared to LLS-based and other SDP-based methods. Notably, when the standard deviation of shading is less than 4dBm, our method performs similarly to ML-true.

\textit{2) Impacts on Communication Range:} In our previous experiments, we assumed a fully connected network. However, in reality, vehicles can only access RSUs within the communication range of the OBU. Fig. \ref{fig8} (b) illustrates how different communication ranges affect ALEs. When the communication range increases from 60m to 120m, CV2X-LOCA's ALE decreases slightly due to increased access to more RSUs for localization information. But, beyond 120m, performance degrades as C-V2X approaches full connectivity. Interestingly, all methods experience an increase in positioning error when the communication range exceeds 120m. This counter-intuitive observation is explained by the fact that while increasing the communication range allows for sharing information with more RSUs, it also dramatically increases inaccurate noise data collection from faraway RSUs. Overall, except for ML-True method, CV2X-LOCA offers superior performance compared to other methods under consideration.

\textit{3) Impacts on Deployment Spacing:} The drawback of using CV2X-based localization methods is that it necessitates constant communication with multiple RSUs. Therefore, the trade-off between positioning accuracy and implementation complexity becomes a crucial factor. In our experiments, we varied the spacing of RSUs from 30m to 210m and observed the positioning accuracy of different methods in Fig. \ref{fig8} (c). As the deployment spacing of RSUs increased, all methods' positioning accuracy decreased to varying degrees. Notably, when the spacing reached 150m, there was a sharp decline in positioning accuracy due to reduced localization feature information resulting in under-fitting of the framework. Conversely, deploying more RSUs can enhance positioning accuracy but also increases computational complexity and deployment costs. Considering these trade-offs, a feasible solution would be deploying RSUs at intervals ranging from 120-150m.

\section{Field Experiment}  \label{V}
\subsection{Field Setup}
\begin{figure}[!t]
\centering
\includegraphics[width=3.4in,height=2in]{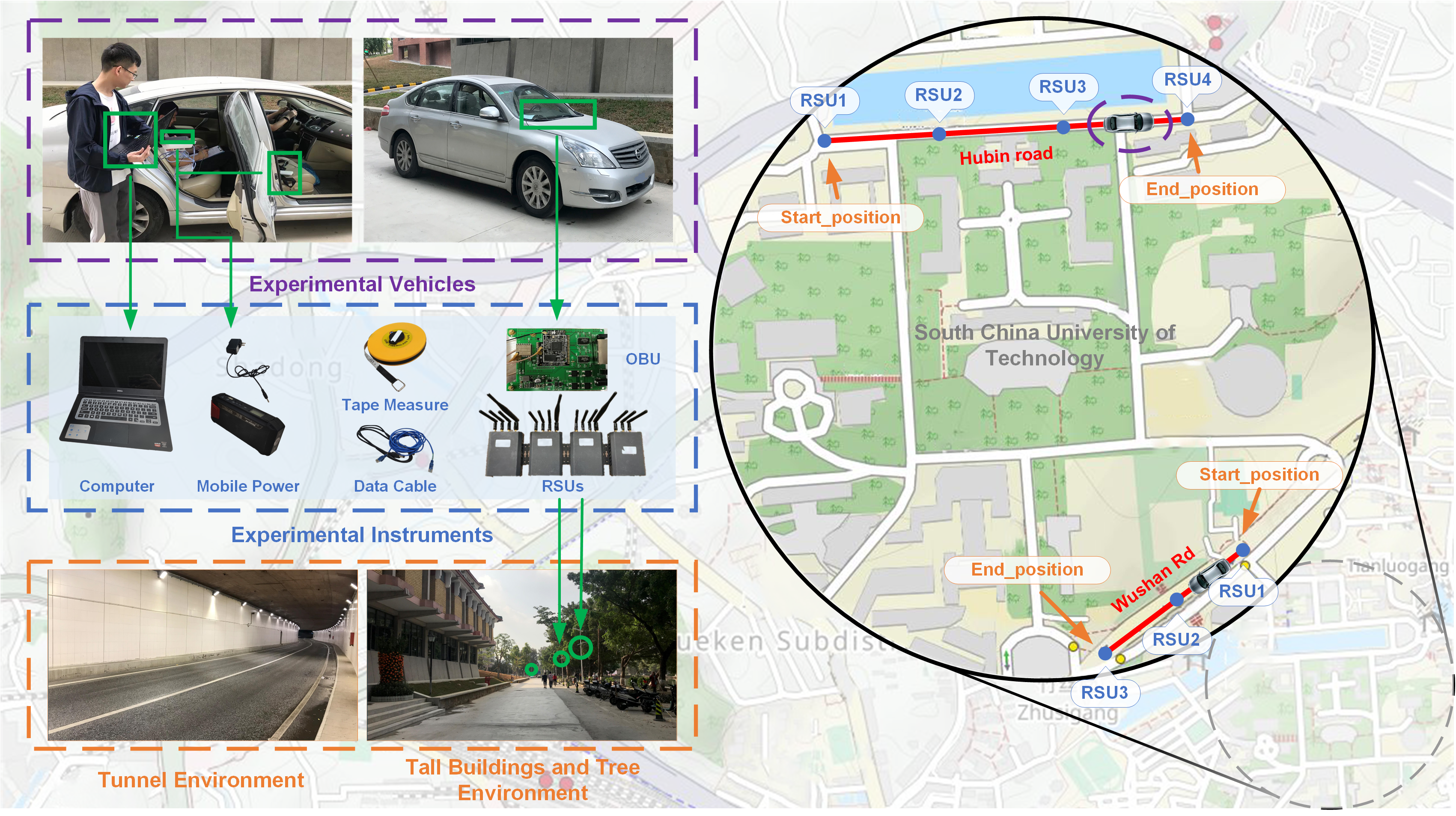}
\caption{The equipment used and the geographical location of field experiment.}
\label{fig9}
\end{figure}

\textit{1) Experiment Scenarios:} This section provides a detailed description and geographical location of the experiment. The first real-world road environment (i) is located on Hubin Road, which is an open road segment in South China University of Technology, Guangzhou, China. As shown in Fig. \ref{fig9}, there is a tall building (Duxing Building with 20 floors) on the left side of the road and a 10-meter-tall tree on the right side. Additionally, other obstacles such as vehicles, pedestrians, bicycles, trucks and metal bars are present around this site leading to frequent interruptions in GNSS signals. The second real-world road environment (ii) is situated in Wushan Road tunnel which is a closed two-way urban road segment located in Tianhe District, Guangzhou, China. Each lane of this tunnel is about 4.25m wide making it a typical GNSS-denied environment for testing purposes. Real tests were conducted at these two locations on January 7th and May 16th, 2021, respectively during which we set the detection interval to be consistent with Li \textit{et al.} 's study \cite{li2018rse} at every 0.1 seconds. After receiving beacon messages from the first RSU installed along each route; vehicles continued moving until they received messages from third RSUs following which OBU clients calculated their coordinates accurately.

\textit{2) Experiment Devices:} Our prototype system, as illustrated in Fig. \ref{fig9}, comprises a notebook computer serving as the data collection server, an OBU, and four RSUs. The OBU and computer were installed near the windshield of our experimental vehicle to collect and process C-V2X signal data. We used an open-source MySQL database to extract standard message messages received from the OBU. The RSUs were deployed on top of temporary test poles or traffic signal poles alongside roadsides. Communication between the OBU, RSUs, and notebook computer was established via LAN or cable. According to Kiela \textit{et al.} \cite{kiela2020review}, Qualcomm Snapdragon 602 and Qualcomm Snapdragon 820 can be utilized as C-V2X devices with long-term evolution (LTE), 4G, 5G, and Wi-Fi communication modules. In this experiment, we employed Qualcomm Snapdragon 820 as our OBU device while using transceiver chipsets manufactured by Chengdu DataSky Company of China for our RSUs as they show excellent performance and support DC power supply with wide voltage range. We used portable power for short-term experiments although installation correction is necessary for all RSUs; coordination of height orientation and antenna direction is not complicated since all RSUs are of the same version without requiring transmission power adjustment.

\textit{3) Ground Truth:} In tunnels, even high-precision GNSS systems can experience significant localization errors. Additionally, manually measuring the location of vehicles while driving is impractical. As a result, obtaining accurate ground truth for comparing different localization methods is challenging. To address this issue, we followed a similar approach to Liu \textit{et al.} \cite{liu2022research}. We took pictures of the surrounding environment (including buildings and landmarks) near RSUs and cross-referenced them with Google Street View to obtain their coordinates on Google Maps accurately. During testing, the vehicle was driven at a uniform speed as much as possible to minimize error. Ground truth was obtained using a laser range finder by comparing depth readings of light poles measured with both the laser range finder and tape measure over distances ranging from 1m to 100m. The resulting error was within 0.15m, which is an acceptable level of accuracy.

\subsection{Performance Comparison}
\begin{figure*}[!t]
\centering
\subfloat[]{\includegraphics[width=0.3\textwidth,height=1.7in]{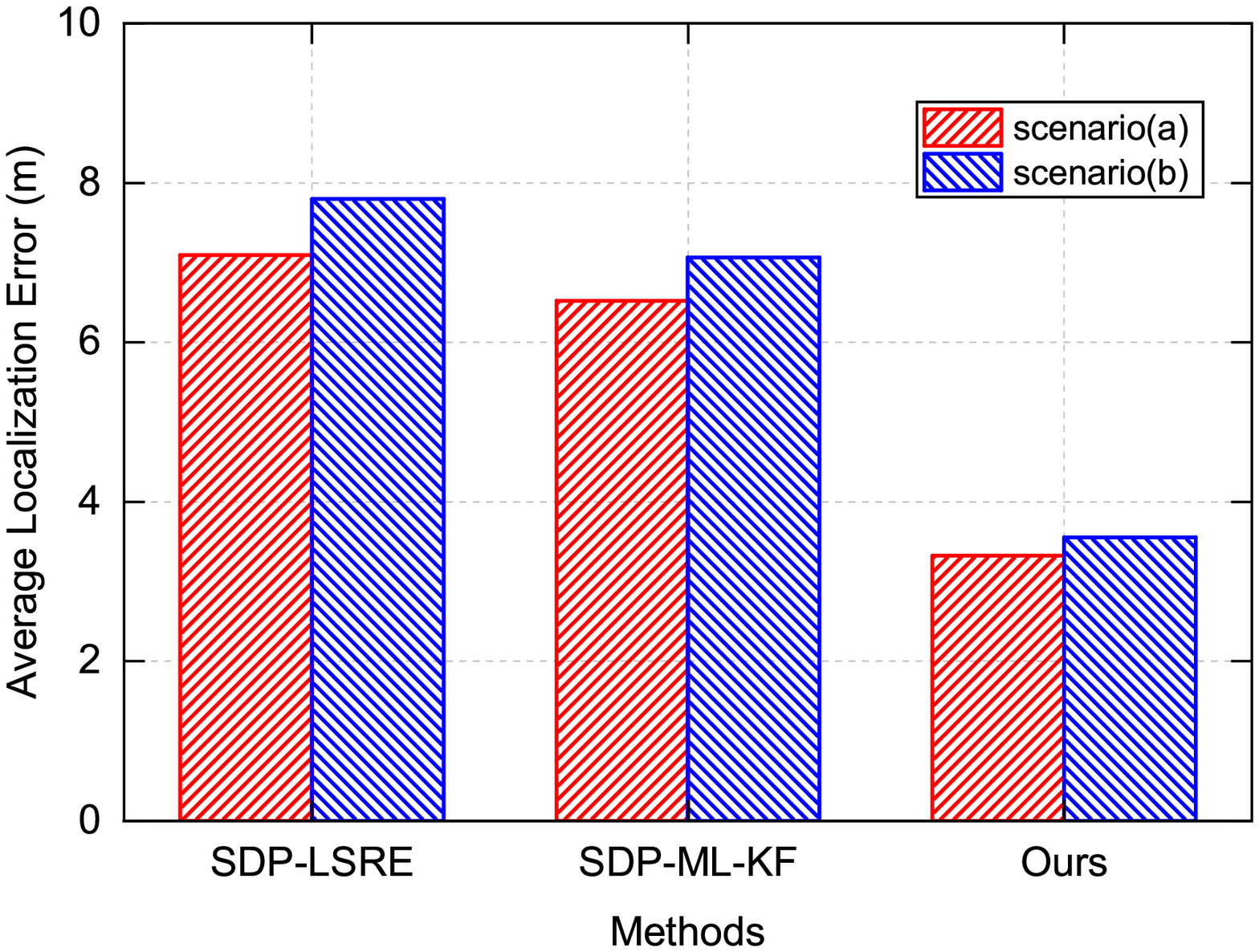}
\label{fig10_1}}
\quad
\subfloat[]{\includegraphics[width=0.3\textwidth,height=1.7in]{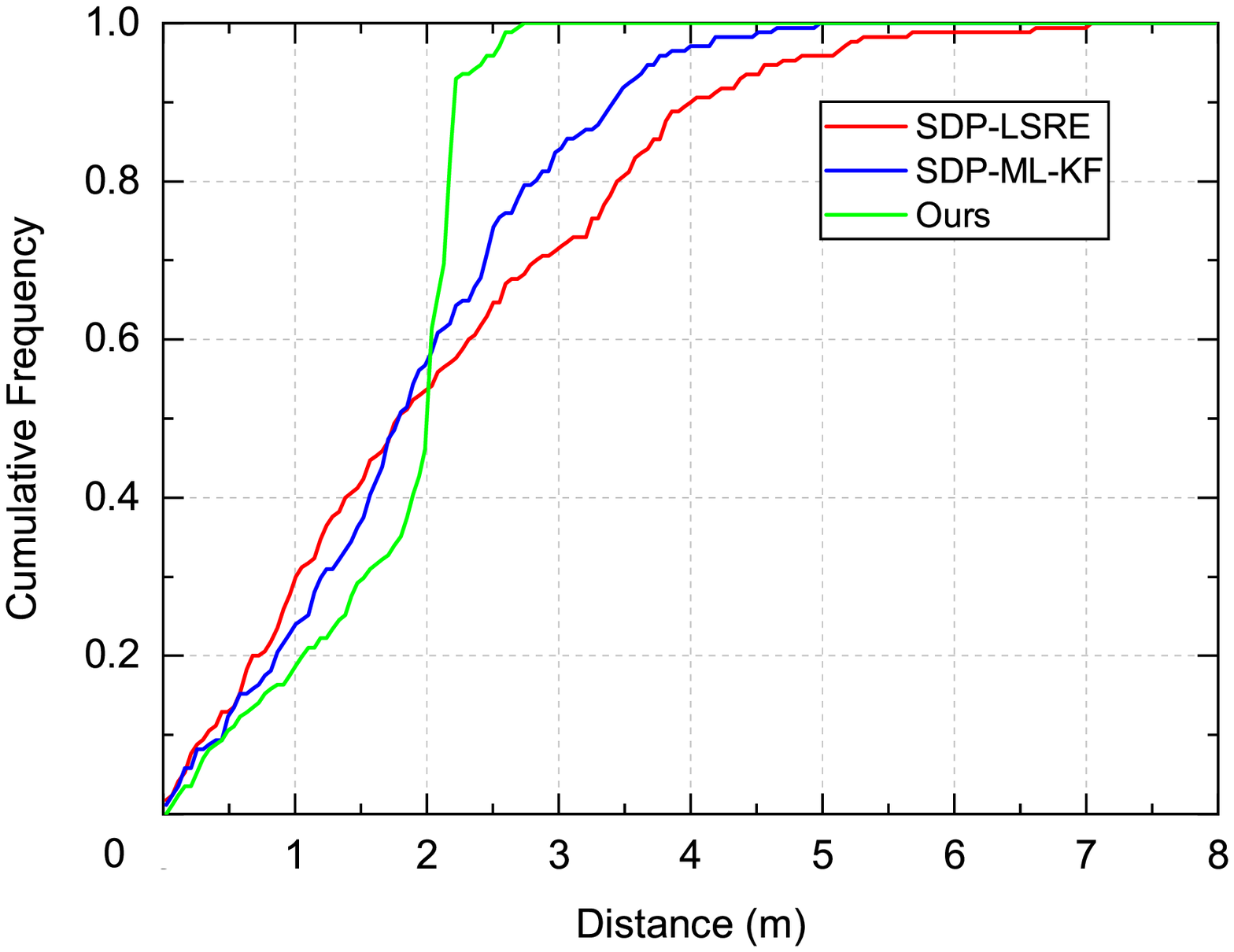}
\label{fig10_2}}
\quad
\subfloat[]{\includegraphics[width=0.3\textwidth,height=1.7in]{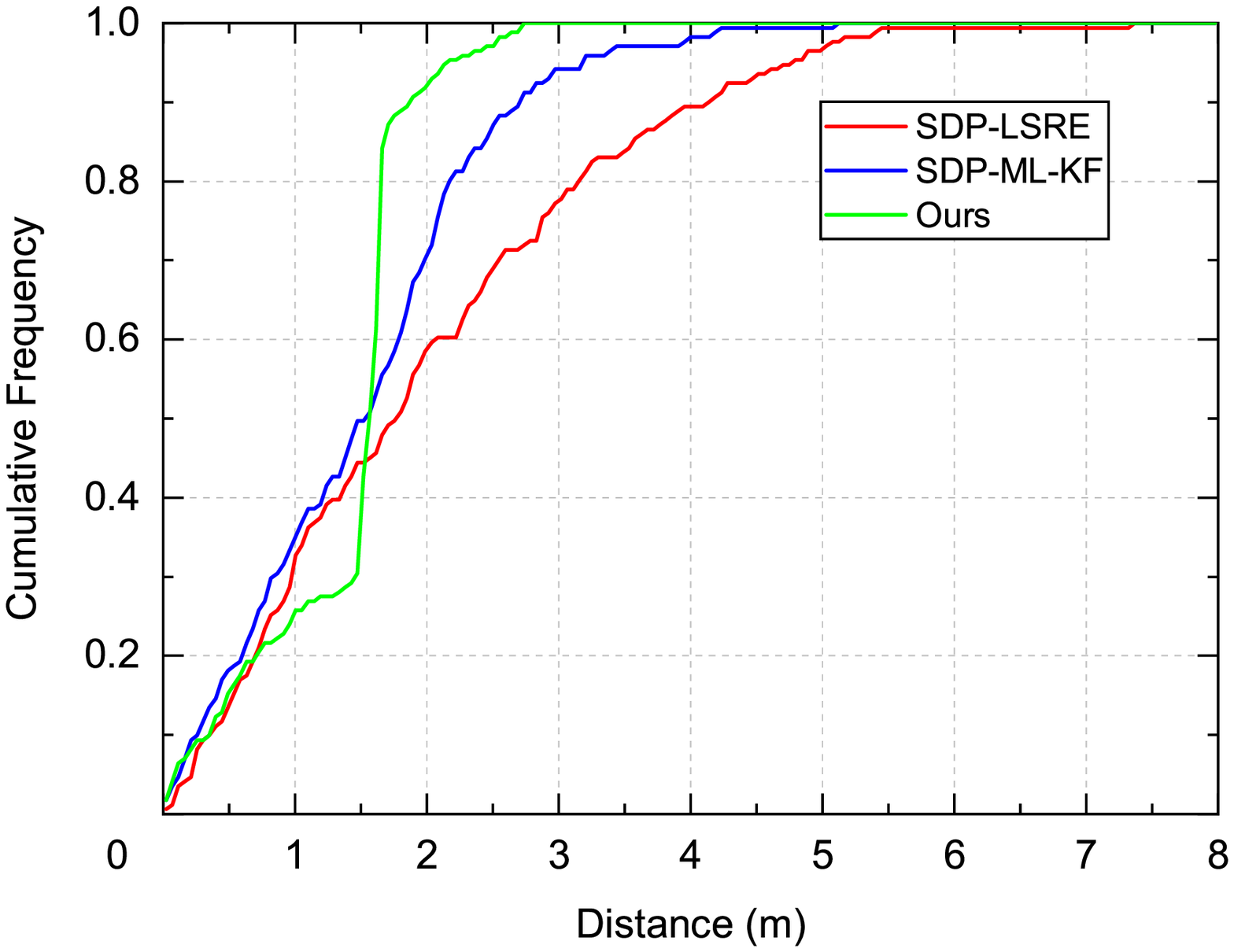}
\label{fig10_3}}
\quad

\subfloat[]{\includegraphics[width=0.3\textwidth,height=1.7in]{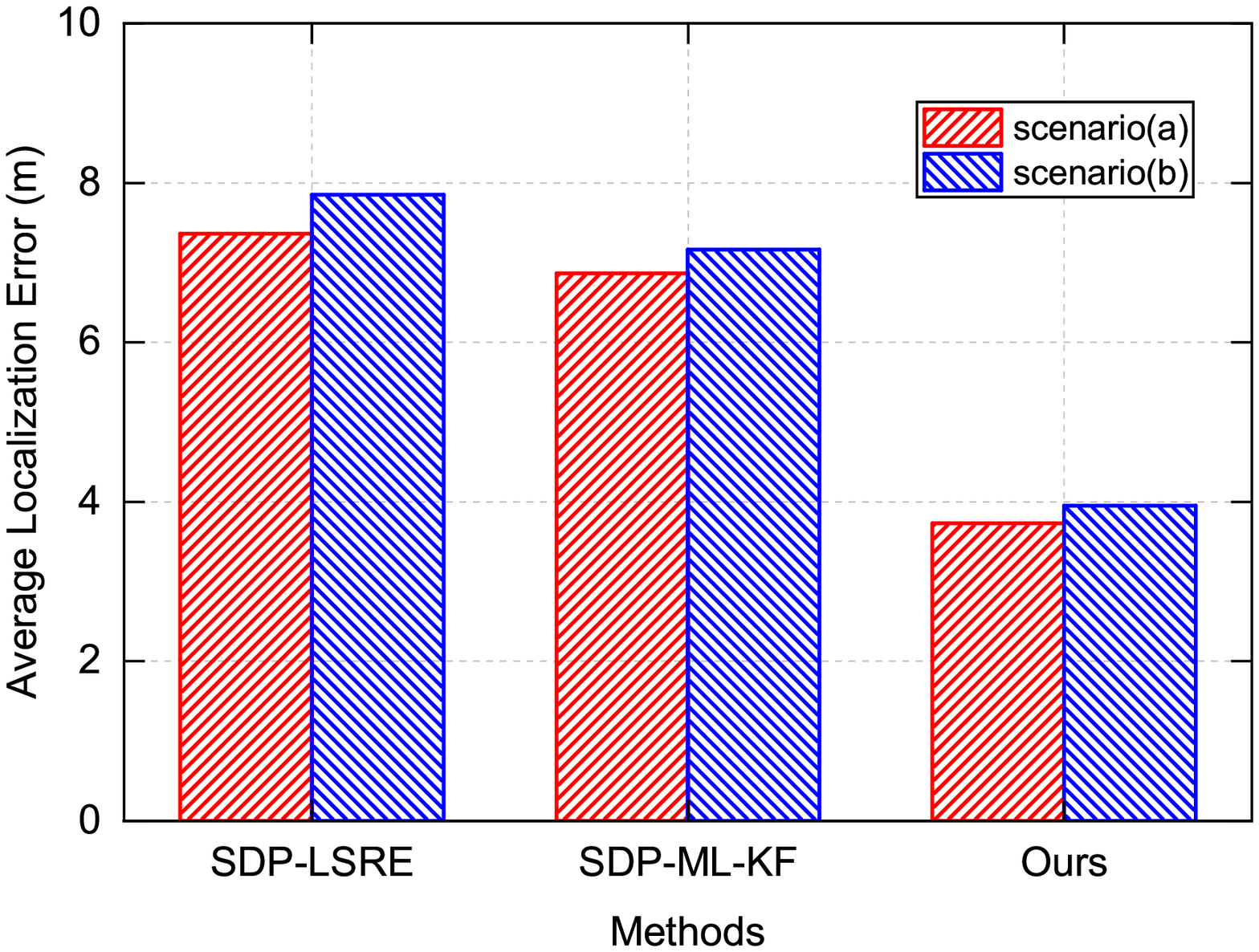}
\label{fig10_4}}
\quad
\subfloat[]{\includegraphics[width=0.3\textwidth,height=1.7in]{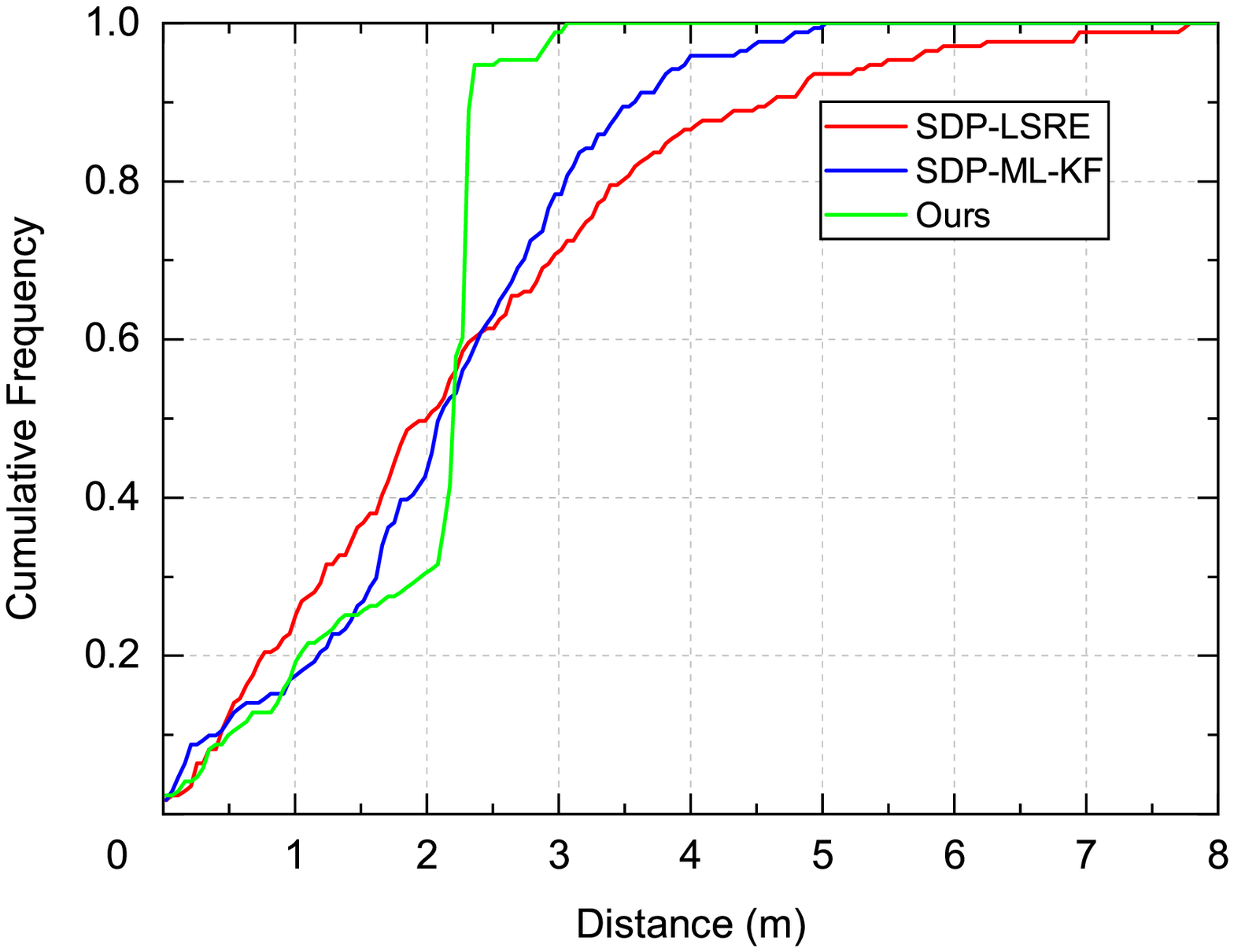}
\label{fig10_5}}
\quad
\subfloat[]{\includegraphics[width=0.3\textwidth,height=1.7in]{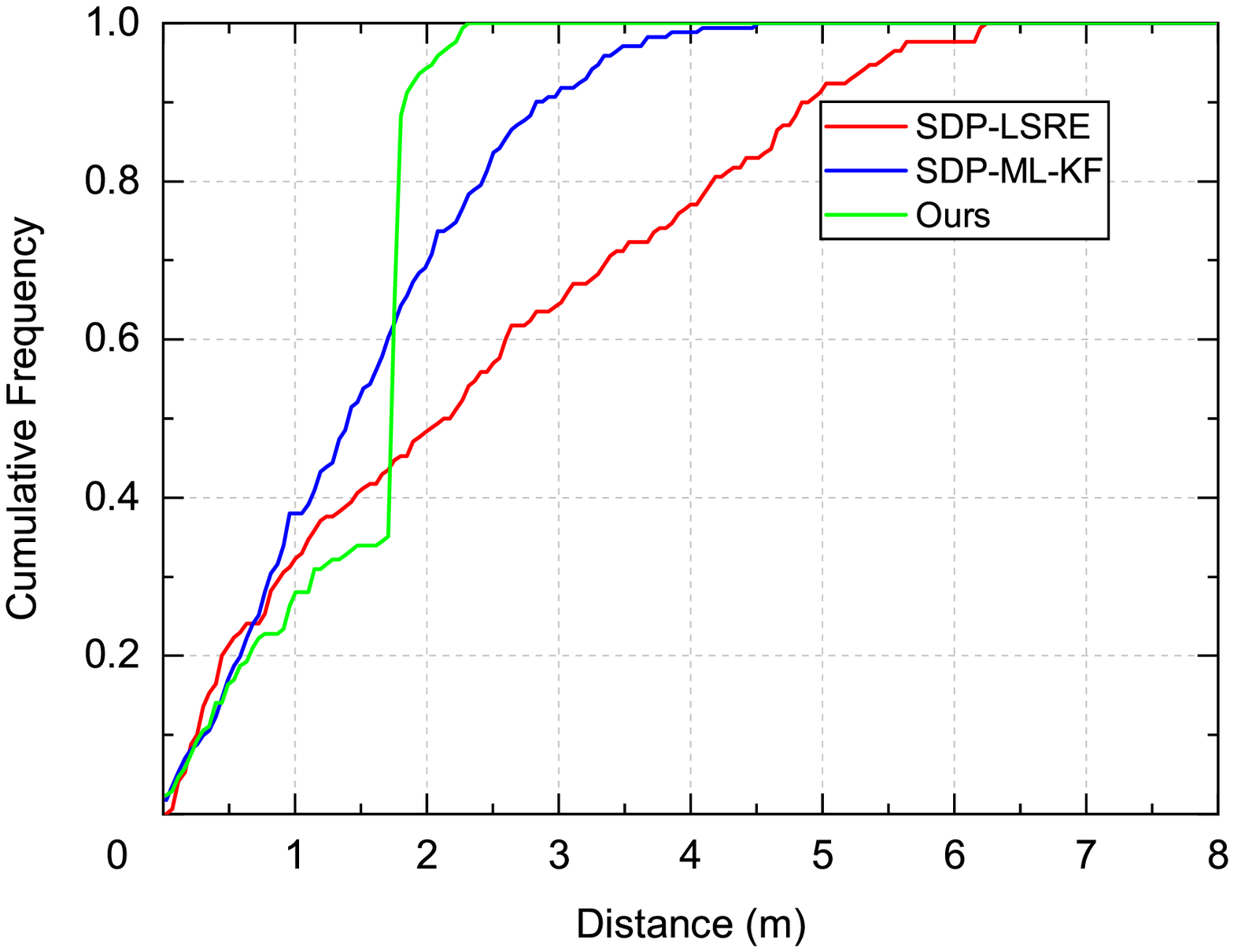}
\label{fig10_6}}
\caption{The localization performance of different methods under real-world road environments (i) and (ii). (a) the change of ALEs under two real-world road environments (vehicle speed=25km/h). (b) $\sim$ (c) the CDFs of longitudinal coordinates under real-world road environments (i) and (ii), respectively (vehicle speed=25km/h). (d) the change of ALEs under two real-world road environments (vehicle speed=60km/h). (e) $\sim$ (f) the CDFs of longitudinal coordinates under real-world road environments (i) and (ii), respectively (vehicle speed=60km/h).}
\label{fig10}
\end{figure*}
This section examines the localization performance of various real-world road environments at different speeds (25km/h and 60km/h). In real-world road environment (i), the test vehicle transitions from lane 1 to lane 2, while in real-world road environment (ii), it drives in lanes 1 or 2 at default speed. Simulation results indicate that SDP-based frameworks, such as SDP-LSRE, SDP-ML-KF, and CV2X-LOCA, exhibit superior positioning accuracy and robustness. To provide readers with a more convenient analysis of these methods' localization performance during field experiments, we conducted multiple repetitions of tests for both road environments at speeds of 25 km/h and 60 km/h (the urban road speed limit). The field test results for each method are presented in Fig. \ref{fig10} (a) $\sim$ (f), with ALE values shown in Fig. \ref{fig10} (a) and (d). Notably, ALE calculates errors longitudinally and laterally. We also analyzed longitudinal error - which is particularly relevant to researchers and traffic engineers - using cumulative distribution functions displayed in Fig. \ref{fig10} (b), (c), (e), and (f).

In the first field test at a speed of 25 km/h, CV2X-LOCA demonstrates superior localization performance in both real-world road environments (i) and (ii), with an ALE of 3.33m and 3.56m respectively, as well as achieving a 90th percentile error of only 2.3m and 2.1m in those same environments. In comparison, the SDP-LSRE achieves a 90th percentile error of 4.5m and 5.1m, while the SDP-ML-KF achieves a 90th percentile error of 3.4m and 2.8m, respectively. In the second field test conducted at a speed of 60 km/h, CV2X-LOCA again outperforms its competitors by achieving an ALE of of 3.73m and 3.96m for both real-world road environments (i) and (ii). Additionally, CV2X-LOCA achieves a 90th percentile error of 2.34m and 1.84m in real-world road environments (i) and (ii) - significantly lower than that achieved by either SDP-LSRE (4.63m and 4.93m) or SDP-ML-KF (2.73m and 2.84m). To summarize, CV2X-LOCA maintains the best localization performance in real-world road environments (i) and (ii) regardless of the speed of 25km/h or 60km/h.

Compared to simulated results, all methods exhibit more significant positioning errors in challenging real-world road environments. This reflects the impact of practical implementation, such as variations in antenna orientation maps and inaccurate signal measurements. However, CV2X-LOCA can provide lower 2D coordinate errors for different real-word road environments and different level of speed (both \textless 4m), while longitudinal coordinate errors of CV2X-LOCA are all less than 2.5m (as shown by green line in Fig. \ref{fig10}). As defined in Li \textit{et al.} \cite{li2018rse}, this level of accuracy can be considered lane-level positioning accuracy. These field test results validate the effectiveness of the CV2X-LOCA framework in practical applications.

\section{Computational Complexity}  \label{VI}
Employing the Gauss-Newton method, the complexity of ML-True is $O(K N)^{2}$, where $K$ is the number of iterations. The computational complexity of the WLS method is $O\left(2 m^{2} n+m n^{2}+m n+n^{3}+n^{2}\right)$. Assuming connected vehicles have more than three neighboring RSUs, the complexity of the LLS method is $O\left(4(N)^{2}+6 N+12\right)$ for the $K$-1 iterations. 
The worst-case complexity of solving an SDP is 
\begin{equation}
O\left(K\left(C^{2} \sum_{c=1}^{N_{s d}}\left(n_{c}^{s d}\right)^{2}+C \sum_{c=1}^{N_{s d}}\left(n_{c}^{s d}\right)^{2}+C^{3}\right) \log \frac{1}{\varepsilon} \right)
\label{Eq29}
\end{equation}
where $C$ is the number of equality constraints, $N_{s d}$ is the number of semi-definite constraints, $n_{c}^{s d}$ is the dimension of the $c$th semi-definite cone, and $\varepsilon$ is the SDP solution accuracy. 

The difference between SDP-LSRE, SDP-ML-KF, and CV2X-LOCA is that the $C$ are $2 N+3$, $(N)^{2}+5.5 N+6$ and $(N)^{2}+5.5 N+6$, respectively. In addition, the computational complexity of the EKF is $O\left(D^{2.4}+\left(D^{\prime}\right)^{2}\right)$. In this case, $D$=2, and $D^{\prime}$=4. The computational complexity of UKF and EKF is of the same magnitude, and UKF only has one more constant term. For 100 experimental repetitions, the average computation times for SDP-LSRE, SDP-ML-KF, and CV2X-LOCA are 8.26, 9.42, and 11.28ms, respectively. The proposed framework has a reasonable complexity compared to other methods. Given the significant improvement in localization accuracy of CV2X-LOCA over the other methods, this moderate increase in complexity may be acceptable.

\section{Conclusion}  \label{VII}
In this research, we propose the first RSU-enabled cooperative localization framework based on C-V2X channel state information, namely CV2X-LOCA, for AVs to achieve lane-level positioning accuracy under GNSS-denied environments. Both simulations and field tests have demonstrated that CV2X-LOCA can achieve state-of-the-art localization performance and maintain impressive robustness across different road environments, RSUs deployment settings, and communication environments. Hence, this framework can serve as an alternative method for AVs to achieve lane-level global positioning accuracy when GNSS signals are not available. 

The proposed framework does not necessitate on-board perception sensors in vehicles, which can help decrease the operating and design expenses of AVs. Additionally, the research findings regarding the optimal connectivity range can offer valuable insights to transportation agencies. This includes determining the spacing between RSUs for vehicle-to-infrastructure communication to minimize volatility in network topology, and identifying cost-effective solutions for deploying RSUs to enhance AVs operations.

It is worth considering the impact of shadowing effect on communication signals due to blocking line of sight (LOS) by objects (e.g., trucks on the road). More importantly, the proposed framework can be seamlessly integrated with existing GNSS/INS systems as well as on-board perception sensors and map information to improve positioning accuracy. Therefore, future work will focus on extending the CV2X-LOCA framework to integrate with map information or on-board perception sensors to further enhance positioning accuracy.

\begin{appendices}
\section{Proof of Equation. (\ref{Eq7})}\label{App1}
Since $\left\|\boldsymbol{\theta_{j}}-\boldsymbol{\phi_{i}}\right\|>0$ and the positive scaling of the objective function does not affect the minimizer, Eq. (\ref{Eq6}) can be equivalently written as
\begin{equation}
\hat{\boldsymbol{\theta}}_{j}=\arg \min _{\boldsymbol{\theta_{j}}} \max _{i}\left|\log _{10} \frac{\left\|\boldsymbol{\theta_{j}}-\boldsymbol{\phi_{i}}\right\|^{2}}{\beta_{i, j}^{2}}\right|
\label{Eq30}
\end{equation}

We also noticed that
\begin{align}
&\left|\log _{10} \frac{\left\|\boldsymbol{\theta_{j}}-\boldsymbol{\phi_{i}}\right\|^{2}}{\beta_{i, j}^{2}}\right| \notag\\
&=\max \left(\log _{10} \frac{\left\|\boldsymbol{\theta_{j}}-\boldsymbol{\phi_{i}}\right\|^{2}}{\beta_{i, j}^{2}}, \log _{10} \frac{\beta_{i, j}^{2}}{\left\|\boldsymbol{\theta_{j}}-\boldsymbol{\phi_{i}}\right\|^{2}}\right) \notag\\
&=\log _{10}\left(\max \left(\frac{\left\|\boldsymbol{\theta_{j}}-\boldsymbol{\phi_{i}}\right\|^{2}}{\beta_{i, j}^{2}}, \frac{\beta_{i, j}^{2}}{\left\|\boldsymbol{\theta_{j}}-\boldsymbol{\phi_{i}}\right\|^{2}}\right)\right) 
\label{Eq31}
\end{align}

Therefore, Eq. (\ref{Eq30}) can be rewritten as
\begin{equation}
\hat{\boldsymbol{\theta}_{j}}=\arg \min _{\boldsymbol{\theta_{j}}} \max _{i} \log _{10}\left(\max \left(\frac{\left\|\boldsymbol{\theta_{j}}-\boldsymbol{\phi_{i}}\right\|^{2}}{\beta_{i, j}^{2}}, \frac{\beta_{i, j}^{2}}{\left\|\boldsymbol{\theta_{j}}-\boldsymbol{\phi_{i}}\right\|^{2}}\right)\right)
\label{Eq32}
\end{equation}

As we know, $\log _{10}(x)$ is a strictly monotonically increasing function in its domain $(0,+\infty)$ (there is a one-to-one mapping between $\log _{10}(x)$ and $x$). In other words, when $\log _{10}(x)$ is maximized, $x$ is also maximized. Thus, Eq. (\ref{Eq32}) can be equivalently written as Eq. (\ref{Eq7}) 

\section{Proof of Equation. (\ref{Eq13})}\label{App2}
By introducing an auxiliary variable $\boldsymbol{\mu}=\left[\mu_{1, j}, \ldots, \mu_{N, j}\right]^{1}\left(\boldsymbol{\mu} \in \mathbb{R}^{N}\right)$, Eq. (\ref{Eq9}) can be cast as \cite{boyd2004convex}
\begin{align}
&\left(\hat{\theta}_{j}, \hat{\boldsymbol{u}}\right)=\arg \min _{\theta_{j}, u} f(\boldsymbol{\mu})\notag\\
\text { s.t. } &\frac{\left\|\theta_{j}-\phi_{i}\right\|^{2}}{\beta_{i, j}^{2}} \leq \mu_{i, j}; \frac{\beta_{i, j}^{2}}{\left\|\theta_{j}-\phi_{i}\right\|^{2}} \leq \mu_{i, j} 
\label{Eq33}
\end{align}

Obviously, in the Eq. (\ref{Eq33}), $\mu_{i, j}>0$.

Then, the minimization problem Eq. (\ref{Eq33}) can be reformulated as
\begin{align}
&\left(\hat{\theta}_{j}, \hat{\boldsymbol{u}}\right)=\arg \min _{\theta_{j}, u} f(\boldsymbol{\mu})\notag\\
\text { s.t. } & \left\|\theta_{j}-\phi_{i}\right\|^{2} \leq \beta_{i, j}^{2} \mu_{i, j}; \left\|\theta_{j}-\phi_{i}\right\|^{2} \geq \beta_{i, j}^{2} \mu_{i, j}^{-1} 
\label{Eq34}
\end{align}

We can find that Eq. (\ref{Eq34}) is actually the same as Eq. (\ref{Eq33}), since the constraints in Eq. (\ref{Eq34}) already imply that $\left\|\theta_{j}-\phi_{i}\right\|^{2} \neq 0$ and $\mu_{i, j}>0$. Therefore, it is reasonable for us to derive Eq. (\ref{Eq34}) from Eq. (\ref{Eq33}).

Similarly, the Eq. (\ref{Eq34}) also can be rewritten with an auxiliary variable $\boldsymbol{X}\left(\boldsymbol{X} \in \mathbb{S}^{2}\right)$ as Equation. (\ref{Eq13}).
\end{appendices}

\bibliographystyle{IEEEtran}
\bibliography{bare_jrnl_new_sample4}

\begin{thebibliography}{10}
\providecommand{\url}[1]{#1}
\csname url@samestyle\endcsname
\providecommand{\newblock}{\relax}
\providecommand{\bibinfo}[2]{#2}
\providecommand{\BIBentrySTDinterwordspacing}{\spaceskip=0pt\relax}
\providecommand{\BIBentryALTinterwordstretchfactor}{4}
\providecommand{\BIBentryALTinterwordspacing}{\spaceskip=\fontdimen2\font plus
\BIBentryALTinterwordstretchfactor\fontdimen3\font minus
  \fontdimen4\font\relax}
\providecommand{\BIBforeignlanguage}[2]{{%
\expandafter\ifx\csname l@#1\endcsname\relax
\typeout{** WARNING: IEEEtran.bst: No hyphenation pattern has been}%
\typeout{** loaded for the language `#1'. Using the pattern for}%
\typeout{** the default language instead.}%
\else
\language=\csname l@#1\endcsname
\fi
#2}}
\providecommand{\BIBdecl}{\relax}
\BIBdecl

\bibitem{feng2023nature}
S.~Feng, H.~Sun, X.~Yan, H.~Zhu, Z.~Zou, S.~Shen, and H.~X. Liu, ``Dense
  reinforcement learning for safety validation of autonomous vehicles,''
  \emph{Nature}, vol. 615, p. 620–627, 2023.

\bibitem{eskandarian2019research}
A.~Eskandarian, C.~Wu, and C.~Sun, ``Research advances and challenges of
  autonomous and connected ground vehicles,'' \emph{IEEE Trans. Intell. Transp.
  Syst.}, vol.~22, no.~2, pp. 683--711, 2019.

\bibitem{matin2022impacts}
A.~Matin and H.~Dia, ``Impacts of connected and automated vehicles on road
  safety and efficiency: A systematic literature review,'' \emph{IEEE Trans.
  Intell. Transp. Syst.}, vol.~24, no.~3, pp. 2705--2736, 2022.

\bibitem{drivinglevels}
A.~Driving, ``Levels of driving automation as per sae international standard
  j3016.''

\bibitem{lu2021real}
Y.~Lu, H.~Ma, E.~Smart, and H.~Yu, ``Real-time performance-focused localization
  techniques for autonomous vehicle: A review,'' \emph{IEEE Trans. Intell.
  Transp. Syst.}, vol.~23, no.~7, pp. 6082--6100, 2021.

\bibitem{jing2022integrity}
H.~Jing, Y.~Gao, S.~Shahbeigi, and M.~Dianati, ``Integrity monitoring of
  gnss/ins based positioning systems for autonomous vehicles: State-of-the-art
  and open challenges,'' \emph{IEEE Trans. Intell. Transp. Syst.}, vol.~23,
  no.~9, pp. 14\,166--14\,187, 2022.

\bibitem{chiang2020performance}
K.-W. Chiang, G.-J. Tsai, H.-J. Chu, and N.~El-Sheimy, ``Performance
  enhancement of ins/gnss/refreshed-slam integration for acceptable lane-level
  navigation accuracy,'' \emph{IEEE Trans. Veh. Technol.}, vol.~69, no.~3, pp.
  2463--2476, 2020.

\bibitem{zhu2018gnss}
N.~Zhu, J.~Marais, D.~B{\'e}taille, and M.~Berbineau, ``Gnss position integrity
  in urban environments: A review of literature,'' \emph{IEEE Trans. Intell.
  Transp. Syst.}, vol.~19, no.~9, pp. 2762--2778, 2018.

\bibitem{ma2017radar}
H.~Ma, E.~Smart, A.~Ahmed, and D.~Brown, ``Radar image-based positioning for
  usv under gps denial environment,'' \emph{IEEE Trans. Intell. Transp. Syst.},
  vol.~19, no.~1, pp. 72--80, 2017.

\bibitem{kim2022tunnel}
K.~Kim, J.~Im, and G.~Jee, ``Tunnel facility based vehicle localization in
  highway tunnel using 3d lidar,'' \emph{IEEE Trans. Intell. Transp. Syst.},
  vol.~23, no.~10, pp. 17\,575--17\,583, 2022.

\bibitem{panev2018road}
S.~Panev, F.~Vicente, F.~De~la Torre, and V.~Prinet, ``Road curb detection and
  localization with monocular forward-view vehicle camera,'' \emph{IEEE Trans.
  Intell. Transp. Syst.}, vol.~20, no.~9, pp. 3568--3584, 2018.

\bibitem{wang2022pavement}
Y.~Wang, Y.~Wang, I.~W.-H. Ho, W.~Sheng, and L.~Chen, ``Pavement marking
  incorporated with binary code for accurate localization of autonomous
  vehicles,'' \emph{IEEE Trans. Intell. Transp. Syst.}, vol.~23, no.~11, pp.
  22\,290--22\,300, 2022.

\bibitem{qin2017vehicles}
H.~Qin, Y.~Peng, and W.~Zhang, ``Vehicles on rfid: Error-cognitive vehicle
  localization in gps-less environments,'' \emph{IEEE Trans. Veh. Technol.},
  vol.~66, no.~11, pp. 9943--9957, 2017.

\bibitem{rohani2015novel}
M.~Rohani, D.~Gingras, and D.~Gruyer, ``A novel approach for improved vehicular
  positioning using cooperative map matching and dynamic base station dgps
  concept,'' \emph{IEEE Trans. Intell. Transp. Syst.}, vol.~17, no.~1, pp.
  230--239, 2015.

\bibitem{lee2020fail}
S.~Lee and S.-W. Seo, ``Fail-safe multi-modal localization framework using
  heterogeneous map-matching sources,'' \emph{IEEE Trans. Intell. Transp.
  Syst.}, vol.~23, no.~5, pp. 4008--4020, 2020.

\bibitem{xu2022v2x}
R.~Xu, H.~Xiang, Z.~Tu, X.~Xia, M.-H. Yang, and J.~Ma, ``V2x-vit:
  Vehicle-to-everything cooperative perception with vision transformer,'' in
  \emph{Proc. Eur. Conf. Comput. Vis.}\hskip 1em plus 0.5em minus 0.4em\relax
  Springer, 2022, pp. 107--124.

\bibitem{adegoke2019infrastructure}
J.~Zidane, E.~Kampert, C.~R. Ford, S.~A. Birrell, and M.~D. Higgins,
  ``Infrastructure wi-fi for connected autonomous vehicle positioning: A review
  of the state-of-the-art,'' \emph{Veh. Commun.}, vol.~20, p. 100185, 2019.

\bibitem{liu2022research}
X.~Liu, T.~Yang, H.~Chen, and T.~Z. Qiu, ``Research on roadside unit-assisted
  cooperative positioning method for a connected vehicle environment,''
  \emph{Transp. Res. Record}, vol. 2676, no.~1, pp. 78--90, 2022.

\bibitem{del2019network}
J.~A. del Peral-Rosado, G.~Seco-Granados, S.~Kim, and J.~A. L{\'o}pez-Salcedo,
  ``Network design for accurate vehicle localization,'' \emph{IEEE Trans. Veh.
  Technol.}, vol.~68, no.~5, pp. 4316--4327, 2019.

\bibitem{yang2020multi}
P.~Yang, D.~Duan, C.~Chen, X.~Cheng, and L.~Yang, ``Multi-sensor multi-vehicle
  (msmv) localization and mobility tracking for autonomous driving,''
  \emph{IEEE Trans. Veh. Technol.}, vol.~69, no.~12, pp. 14\,355--14\,364,
  2020.

\bibitem{li2018rse}
J.~Li, J.~Gao, H.~Zhang, and T.~Z. Qiu, ``Rse-assisted lane-level positioning
  method for a connected vehicle environment,'' \emph{IEEE Trans. Intell.
  Transp. Syst.}, vol.~20, no.~7, pp. 2644--2656, 2018.

\bibitem{linrssi}
Y.~Lin, Z.~Huang, P.~Wu, and L.~Xu, ``Rssi positioning method of vehicles in
  tunnels based on semi-supervised extreme learning machine,'' \emph{J. Traffic
  Transp. Eng.}, vol.~21, no.~2, pp. 243--255, 2021.

\bibitem{ko2021v2x}
S.-W. Ko, H.~Chae, K.~Han, S.~Lee, D.-W. Seo, and K.~Huang, ``V2x-based
  vehicular positioning: Opportunities, challenges, and future directions,''
  \emph{IEEE Wirel. Commun.}, vol.~28, no.~2, pp. 144--151, 2021.

\bibitem{saeed2018localization}
N.~Saeed, W.~Ahmad, and D.~M.~S. Bhatti, ``Localization of vehicular ad-hoc
  networks with rss based distance estimation,'' in \emph{Proc. Int. Conf.
  Comput. Math. Eng. Technol.}\hskip 1em plus 0.5em minus 0.4em\relax IEEE,
  2018, pp. 1--6.

\bibitem{brambilla2019augmenting}
M.~Brambilla, M.~Nicoli, G.~Soatti, and F.~Deflorio, ``Augmenting vehicle
  localization by cooperative sensing of the driving environment: Insight on
  data association in urban traffic scenarios,'' \emph{IEEE Trans. Intell.
  Transp. Syst.}, vol.~21, no.~4, pp. 1646--1663, 2019.

\bibitem{zhuang2022novel}
C.~Zhuang, H.~Zhao, S.~Hu, X.~Meng, and W.~Feng, ``A novel gnss fault detection
  and exclusion method for cooperative positioning system,'' \emph{IEEE Trans.
  Veh. Technol.}, pp. 1--16, 2022.

\bibitem{moradi2023dsrc}
E.~Moradi-Pari, D.~Tian, M.~Bahramgiri, S.~Rajab, and S.~Bai, ``Dsrc versus
  lte-v2x: Empirical performance analysis of direct vehicular communication
  technologies,'' \emph{IEEE Trans. Intell. Transp. Syst.}, 2023.

\bibitem{maglogiannis2021experimental}
V.~Maglogiannis, D.~Naudts, S.~Hadiwardoyo, D.~van~den Akker, J.~Marquez-Barja,
  and I.~Moerman, ``Experimental v2x evaluation for c-v2x and its-g5
  technologies in a real-life highway environment,'' \emph{IEEE Trans. Netw.
  Serv. Manag.}, vol.~19, no.~2, pp. 1521--1538, 2021.

\bibitem{liu2021highly}
Q.~Liu, P.~Liang, J.~Xia, T.~Wang, M.~Song, X.~Xu, J.~Zhang, Y.~Fan, and
  L.~Liu, ``A highly accurate positioning solution for c-v2x systems,''
  \emph{Sensors}, vol.~21, no.~4, p. 1175, 2021.

\bibitem{kutila2019c}
M.~Kutila, P.~Pyykonen, Q.~Huang, W.~Deng, W.~Lei, and E.~Pollakis, ``C-v2x
  supported automated driving,'' in \emph{Proc. IEEE Int. Conf. Commun.
  Workshops (ICC Workshops)}.\hskip 1em plus 0.5em minus 0.4em\relax IEEE,
  2019, pp. 1--5.

\bibitem{jianyong2014rssi}
Z.~Jianyong, L.~Haiyong, C.~Zili, and L.~Zhaohui, ``Rssi based bluetooth low
  energy indoor positioning,'' in \emph{Proc. IEEE Int. Conf. Indoor
  Positioning Indoor Navigat. (IPIN)}.\hskip 1em plus 0.5em minus 0.4em\relax
  IEEE, 2014, pp. 526--533.

\bibitem{ma2019efficient}
S.~Ma, F.~Wen, X.~Zhao, Z.-M. Wang, and D.~Yang, ``An efficient v2x based
  vehicle localization using single rsu and single receiver,'' \emph{IEEE
  Access}, vol.~7, pp. 46\,114--46\,121, 2019.

\bibitem{chen2016vehicle}
C.-H. Chen, C.-A. Lee, and C.-C. Lo, ``Vehicle localization and velocity
  estimation based on mobile phone sensing,'' \emph{IEEE Access}, vol.~4, pp.
  803--817, 2016.

\bibitem{magowe2019closed}
K.~Magowe, A.~Giorgetti, and K.~Sithamparanathan, ``Closed-form approximation
  of weighted centroid localization performance,'' \emph{IEEE Sensors Letters},
  vol.~3, no.~12, pp. 1--4, 2019.

\bibitem{zheng2020accurate}
R.~Zheng, G.~Wang, and K.~Ho, ``Accurate semidefinite relaxation method for
  elliptic localization with unknown transmitter position,'' \emph{IEEE Trans.
  Wirel. Commun.}, vol.~20, no.~4, pp. 2746--2760, 2020.

\bibitem{wang2018cooperative}
Z.~Wang, H.~Zhang, T.~Lu, and T.~A. Gulliver, ``Cooperative rss-based
  localization in wireless sensor networks using relative error estimation and
  semidefinite programming,'' \emph{IEEE Trans. Veh. Technol.}, vol.~68, no.~1,
  pp. 483--497, 2018.

\bibitem{zhang2021localization}
Y.~Zhang, X.~Gong, K.~Liu, and S.~Zhang, ``Localization and tracking of an
  indoor autonomous vehicle based on the phase difference of passive uhf rfid
  signals,'' \emph{Sensors}, vol.~21, no.~9, p. 3286, 2021.

\bibitem{zou2021rss}
Y.~Zou and H.~Liu, ``Rss-based target localization with unknown model
  parameters and sensor position errors,'' \emph{IEEE Trans. Veh. Technol.},
  vol.~70, no.~7, pp. 6969--6982, 2021.

\bibitem{page2019enhanced}
M.~Page and T.~L. Wickramarathne, ``Enhanced situational awareness with signals
  of opportunity: Rss-based localization and tracking,'' in \emph{Proc. IEEE
  Intell. Transp. Syst. Conf. (ITSC)}.\hskip 1em plus 0.5em minus 0.4em\relax
  IEEE, 2019, pp. 3833--3838.

\bibitem{huang2020multi}
Z.~Huang, L.~Xu, and Y.~Lin, ``Multi-stage pedestrian positioning using
  filtered wifi scanner data in an urban road environment,'' \emph{Sensors},
  vol.~20, no.~11, p. 3259, 2020.

\bibitem{jondhale2018kalman}
S.~R. Jondhale and R.~S. Deshpande, ``Kalman filtering framework-based real
  time target tracking in wireless sensor networks using generalized regression
  neural networks,'' \emph{IEEE Sens. J.}, vol.~19, no.~1, pp. 224--233, 2018.

\bibitem{soto2022survey}
I.~Soto, M.~Calderon, O.~Amador, and M.~Urue{\~n}a, ``A survey on road safety
  and traffic efficiency vehicular applications based on c-v2x technologies,''
  \emph{Veh. Commun.}, vol.~33, p. 100428, 2022.

\bibitem{mafakheri2021optimizations}
B.~Mafakheri, P.~Gonnella, A.~Bazzi, B.~M. Masini, M.~Caggiano, and R.~Verdone,
  ``Optimizations for hardware-in-the-loop-based v2x validation platforms,'' in
  \emph{Proc. IEEE 93rd Veh. Technol. Conf.}\hskip 1em plus 0.5em minus
  0.4em\relax IEEE, 2021, pp. 1--7.

\bibitem{miao2022does}
L.~Miao, S.-F. Chen, Y.-L. Hsu, and K.-L. Hua, ``How does c-v2x help autonomous
  driving to avoid accidents?'' \emph{Sensors}, vol.~22, no.~2, p. 686, 2022.

\bibitem{gholami2011positioning}
M.~R. Gholami, \emph{Positioning algorithms for wireless sensor
  networks}.\hskip 1em plus 0.5em minus 0.4em\relax Chalmers Tekniska Hogskola
  (Sweden), 2011.

\bibitem{ouyang2010received}
R.~W. Ouyang, A.~K.-S. Wong, and C.-T. Lea, ``Received signal strength-based
  wireless localization via semidefinite programming: Noncooperative and
  cooperative schemes,'' \emph{IEEE Trans. Veh. Technol.}, vol.~59, no.~3, pp.
  1307--1318, 2010.

\bibitem{huang2019novel}
Z.~Huang, X.~Zhu, Y.~Lin, L.~Xu, and Y.~Mao, ``A novel wifi-oriented rssi
  signal processing method for tracking low-speed pedestrians,'' in \emph{Proc.
  5th Int. Conf. Transp. Inf. Saf. (ICTIS)}.\hskip 1em plus 0.5em minus
  0.4em\relax IEEE, 2019, pp. 1018--1023.

\bibitem{nguyen2022cellular}
H.~Nguyen, M.~Noor-A-Rahim, Y.~L. Guan, and D.~Pesch, ``Cellular v2x
  communications in the presence of big vehicle shadowing: Performance analysis
  and mitigation,'' \emph{IEEE Trans. Veh. Technol.}, vol.~72, no.~3, pp.
  3764--3776, 2022.

\bibitem{jo2016tracking}
K.~Jo, M.~Lee, J.~Kim, and M.~Sunwoo, ``Tracking and behavior reasoning of
  moving vehicles based on roadway geometry constraints,'' \emph{IEEE Trans.
  Intell. Transp. Syst.}, vol.~18, no.~2, pp. 460--476, 2016.

\bibitem{etsi2011intelligent}
T.~ETSI, ``Intelligent transport systems (its); vehicular communications; basic
  set of applications; part 2: Specification of cooperative awareness basic
  service,'' \emph{Draft ETSI TS}, vol.~20, no. 2011, pp. 448--51, 2011.

\bibitem{sae2022V2X}
SAE, ``V2x communications message set dictionary j2735\_202211,'' \emph{V2X
  Core Technical Committee}, vol.~20, Nov 2022.

\bibitem{boyd2004convex}
S.~Boyd, S.~P. Boyd, and L.~Vandenberghe, \emph{Convex optimization}.\hskip 1em
  plus 0.5em minus 0.4em\relax Cambridge univ. press, 2004.

\bibitem{boyd1994linear}
S.~Boyd, L.~El~Ghaoui, E.~Feron, and V.~Balakrishnan, \emph{Linear matrix
  inequalities in system and control theory}.\hskip 1em plus 0.5em minus
  0.4em\relax SIAM, 1994.

\bibitem{sturm1999using}
J.~F. Sturm, ``Using sedumi 1.02, a matlab toolbox for optimization over
  symmetric cones,'' \emph{Optim. Method Softw.}, vol.~11, no. 1-4, pp.
  625--653, 1999.

\bibitem{gibreel1999state}
G.~Gibreel, S.~Easa, Y.~Hassan, and I.~El-Dimeery, ``State of the art of
  highway geometric design consistency,'' \emph{J. Transp. Eng.}, vol. 125,
  no.~4, pp. 305--313, 1999.

\bibitem{klapez2020application}
M.~Klapez, C.~A. Grazia, and M.~Casoni, ``Application-level performance of ieee
  802.11 p in safety-related v2x field trials,'' \emph{IEEE Internet Things
  J.}, vol.~7, no.~5, pp. 3850--3860, 2020.

\bibitem{kiela2020review}
K.~Kiela, V.~Barzdenas, M.~Jurgo, V.~Macaitis, J.~Rafanavicius, A.~Vasjanov,
  L.~Kladovscikov, and R.~Navickas, ``Review of v2x--iot standards and
  frameworks for its applications,'' \emph{Appl. sci.}, vol.~10, no.~12, p.
  4314, 2020.

\end{thebibliography}

\vfill

\end{document}